\documentclass[10pt,twocolumn,letterpaper]{article}

\usepackage[pagenumbers]{cvpr}
\usepackage{placeins}

\usepackage{multirow}
\usepackage{colortbl}
\definecolor{oursrow}{RGB}{228,230,250}
\newcommand{\best}[1]{\textbf{#1}}
\newcommand{\second}[1]{\underline{#1}}
\newcommand{\tworow}[2]{\begin{tabular}[c]{@{}c@{}}#1\\#2\end{tabular}}
\newcommand{\up}[1]{\textsubscript{\textcolor{red}{$\uparrow$#1}}}
\newcommand{\down}[1]{\textsubscript{\textcolor{green}{$\downarrow$#1}}}

\newcommand{\papertitle}{Spherical Multi-Prototype Learning for Generalizable AI-Generated Image Detection}

\definecolor{cvprblue}{rgb}{0.21,0.49,0.74}
\usepackage[breaklinks,colorlinks,allcolors=cvprblue]{hyperref}
\hypersetup{
  pdftitle={\papertitle},
  pdfauthor={Kun Guo, Haoyue Wang, Sheng Li, Xinpeng Zhang, Zhenxing Qian}
}

\newcommand{\abstractcodeavailability}{The code will be available at \url{https://github.com/guokun111/PE-SMP}.}

\title{\papertitle}
\author{%
  \textbf{Kun Guo}\textsuperscript{1},
  \textbf{Haoyue Wang}\textsuperscript{1},
  \textbf{Sheng Li}\textsuperscript{1},
  \textbf{Xinpeng Zhang}\textsuperscript{1},
  \textbf{Zhenxing Qian}\textsuperscript{1*}\\
  {\normalsize
   \textsuperscript{1}\,College of Computer Science and Artificial Intelligence, Fudan University
   \quad \textsuperscript{*}\,Corresponding author}
}

\begin{document}
\maketitle
\providecommand{\abstractcodeavailability}{The code will be made publicly available after publication.}

\begin{abstract}
AI-generated image detection requires generalization and robustness to unseen scenarios.
Many existing detectors rely on low-level forensic traces that common post-processing can weaken or erase.
Some detectors draw on pretrained vision models to distinguish real from generated images.
However, such methods may underuse the rich semantic structure of the learned representations, limiting generalization to unseen scenarios.
To address this challenge, we propose PE-SMP, which pairs a frozen Perception Encoder(PE) with a Spherical Multi-Prototype (SMP) head.
Its design is guided by two observations of PE features.
First, real and generated feature centroids exhibit systematic angular offsets within semantic classes.
Second, these offsets structure exhibit a shared direction, with residual directions that align more closely among semantically related classes.
To exploit this structure in pretrained features, we design SMP to learn paired real and fake prototypes for each visual context using shared and local offsets.
It aggregates angular distance differences across prototype pairs through soft routing.
Extensive experiments show that PE-SMP, trained solely on GenImage SD v1.4, outperforms prior baselines by 23.88\% across seven benchmarks and remains robust to common post-processing.
SMP also improves mean accuracy in all five tested pretrained backbone configurations.
\abstractcodeavailability
\end{abstract}

\section{Introduction}
\label{sec:intro}

\begin{figure}[t]
  \centering
  \includegraphics[width=\linewidth]{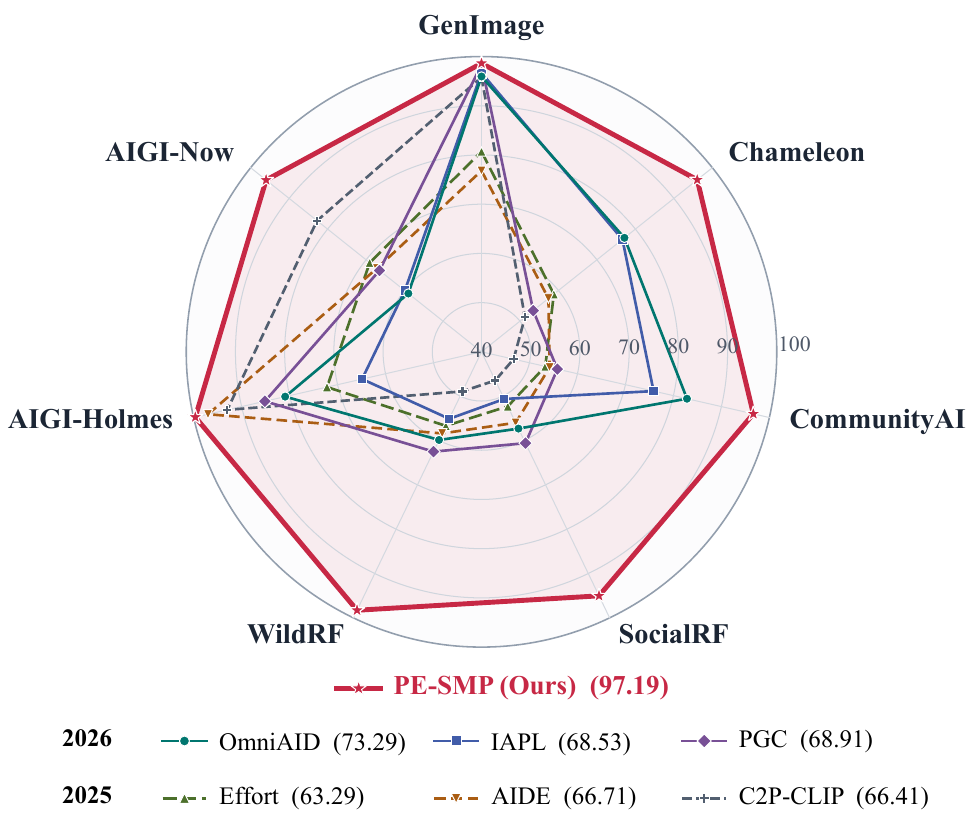}
   \caption{Cross-dataset generalization of our PE-SMP and six baselines on
    seven test datasets. Each axis shows the accuracy (ACC, \%) on one
    dataset. Values in parentheses are the unweighted means across the
    seven datasets.}
    \label{fig:radar}
\end{figure}

Image generation has advanced rapidly, from generative adversarial networks~\cite{karras2017progressiveprogan,karras2019stylegan} through diffusion models~\cite{rombach2022highLDM,ramesh2022DALL-E3} to recent autoregressive models~\cite{han2025infinity,wu2025janus}. These advances have made realistic images easier to generate at scale.The widespread circulation of these images raises serious concerns about misinformation, fraud, and the credibility of visual media, making AI-generated image detection an increasingly important problem. In practice, images may come from generators unseen during training and may undergo post-processing operations such as compression and resizing when shared on social media. Consequently, generalization and robustness are essential for AI-generated image detection in open-world settings.

Several methods have achieved progressive detection performance by exploiting low-level forensic cues, including reconstruction errors~\cite{wang2023dire,chen2024drct}, local pixel correlations~\cite{tan2024rethinkingNPR}, and upsampling artifacts~\cite{karageorgiou2025Spec_SPAL,frank2020FreDect,tan2024frequencyFreqNet}. These traces can also be generator-specific, making them less reliable for detecting images produced by other architectures.
Other approaches draw on visual priors from pretrained models such as CLIP~\cite{radford2021learningclip,ojha2023towards} and DINOv3~\cite{zhou2026simplicity}. Zhou et al. showed that a linear classifier on frozen Perception Encoder (PE)~\cite{bolya2026perceptionPEcore} features can achieve strong detection performance. However, existing methods may underuse the semantic structure of pretrained features, limiting transfer to images with unfamiliar content.

To address this challenge, we propose PE-SMP, an AI-generated image detector that pairs a frozen Perception Encoder (PE)~\cite{bolya2026perceptionPEcore} with a Spherical Multi-Prototype (SMP) head.
The head uses the semantic structure of pretrained features to compare an image with real and fake prototypes in similar visual contexts.
Its design is motivated by two observations.

\begin{figure}[t]
  \centering
  \includegraphics[width=\columnwidth]{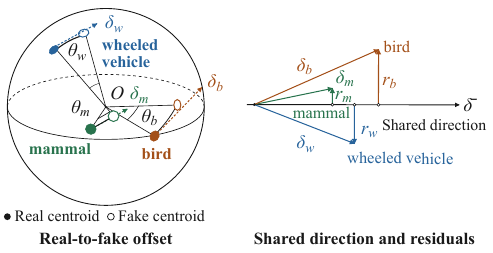}
\caption{\textbf{Class-wise real-to-fake offsets and their decomposition.}
\textbf{Left.} For class $c$, $\delta_c$ is tangent to the unit sphere at
the real centroid (filled) and points in the initial direction of the
shortest arc to the fake centroid (open). Its length is the angle
$\theta_c$ between the centroids.
\textbf{Right.} Each offset is drawn from a common origin and decomposed
into its projection along the mean offset $\bar{\delta}$ (the shared
direction) and a orthogonal class-specific residual $r_c$.}
  \label{fig:smp-offset-definition}
\end{figure}

\begin{figure}[!t]
  \centering
  \includegraphics[width=\linewidth]{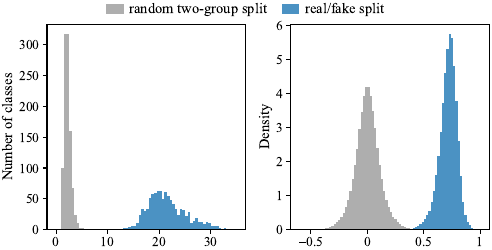}\par
  \subcaptionbox{Within-class offset angle \label{fig:smp-offset-angle}}[0.5\linewidth]{}%
  \subcaptionbox{Cross-class offset cosine\label{fig:smp-offset-cosine}}[0.5\linewidth]{}
\caption{\textbf{Distributions of real-to-fake offset magnitudes and
directional similarities.}
We analyze PE-Core G/14 features from 1,000 ImageNet classes in the
GenImage SD v1.4 training split.
\textbf{(a)} Number of classes in each bin of the angle between real and
fake feature centroids. This angle measures offset magnitude.
\textbf{(b)} Density of cosine similarities between offset directions
for pairs of distinct classes.
Blue uses the real/fake split. Gray uses size-matched random splits
within each class.}
  \label{fig:smp-offset-statistics}
\end{figure}

\begin{figure*}[!t]
  \centering
  \begin{minipage}[t]{\columnwidth}
    \centering
    \vspace{0pt}
    \includegraphics[width=\linewidth]{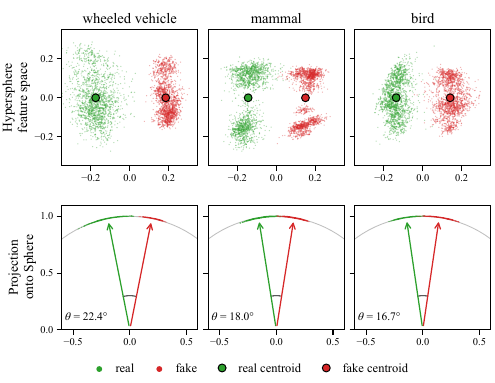}
\caption{\textbf{Real-to-fake offsets in three semantic groups.}
\textbf{Top.} Small dots show projected PE-Core G/14 features of real
(green) and fake (red) images. Large circles mark their centroids.
The horizontal axis points from the real centroid toward the fake centroid.
\textbf{Bottom.} Each cross-section passes through the sphere center and
the two centroids shown above. The angle $\theta$ between the centroid
directions gives the offset magnitude.}
    \label{fig:smp-class-projection}
  \end{minipage}\hfill%
  \begin{minipage}[t]{\columnwidth}
    \centering
    \vspace{0pt}
    \includegraphics[width=\linewidth]{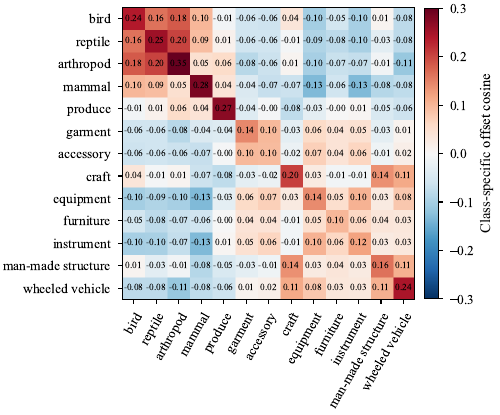}
  \caption{\textbf{Similarity of residual offset directions within and across
semantic groups.}
We remove the shared directional component from each class's real-to-fake
offset. Each cell shows the mean cosine similarity between the remaining
directions, estimated from separate image halves.
Diagonal cells compare distinct classes within the same group, and
off-diagonal cells compare classes from different groups.
Warmer colors indicate more similar directions.}
    \label{fig:smp-offset-residuals}
  \end{minipage}
\end{figure*}

First, inspired by SphereVideo~\cite{li2026spherevideo}, we investigate feature direction as a potential cue for robust AI-generated image detection. We normalize PE-Core features to unit length and examine their angular structure on the hypersphere. For each of the 1,000 semantic classes in the GenImage SD v1.4 training split~\cite{zhu2023genimage}, we define the real-to-fake offset as the displacement on the hypersphere from the normalized centroid of real features to that of generated features. The angle between the centroids gives the magnitude of this offset (\cref{fig:smp-offset-definition}, left). Across the 1,000 classes, the observed angles are systematically larger than those obtained from random within-class splits (\cref{fig:smp-offset-angle}). \Cref{fig:smp-class-projection} illustrates these offsets for wheeled vehicles, mammals, and birds. The results indicate that angular offsets between real and generated features are widespread across semantic classes. This systematic separation suggests that feature direction can help distinguish real from generated images within the same semantic class.

Second, we examine how the directions of real-to-fake offsets relate across classes. We decompose each offset into a component along the average offset direction and an orthogonal residual (\cref{fig:smp-offset-definition}, right). The cosine similarities between offsets from different classes tend to be higher than those obtained from random within-class splits (\cref{fig:smp-offset-cosine}). This suggests that the offsets share a directional component that captures a recurring shift from real to generated features across different types of image content. After removing this shared component, we find that residual directions are more closely aligned for pairs of classes within the same semantic group, such as different bird classes, than for pairs from different groups (\cref{fig:smp-offset-residuals}). This suggests that the remaining shift is associated with what the images depict, motivating the use of local offsets for different visual contexts.

Motivated by these two observations, we design a Spherical Multi-Prototype(SMP) detection head for frozen pretrained vision backbones (\cref{fig:smp-overview}). SMP clusters the training features into visual contexts. For each context, it learns a pair of real and fake prototypes around the cluster centroid by combining a shared offset with a local offset. A bounded adapter refines the feature used for scoring. To score an image, SMP obtains soft routing weights by applying a softmax to the scaled cosine similarities between the original feature and the cluster centroids. Within each context, SMP computes the adapted feature's angular distance to the real prototype minus its angular distance to the fake prototype. The final score is the weighted sum of these differences across contexts. Trained solely on the SD v1.4 subset of GenImage, PE-SMP achieves an average accuracy of 97.19\% across seven benchmarks covering 40 test subsets (\cref{fig:radar}). Our contributions are as follows:
\begin{itemize}
    \item We characterize class-wise real-to-fake offsets in pretrained model features, identifying a shared direction and class-specific residuals associated with image content.
    \item We propose PE-SMP, an AI-generated image detector that pairs a frozen Perception Encoder(PE) with a Spherical Multi-Prototype(SMP) head to score images against real and fake prototypes within visual contexts.
    \item We show that SMP is effective across all five tested pretrained backbone configurations, including models outside the Perception Encoder family.
\end{itemize}

\section{Related work}

\subsection{AI-generated image detection}

Some studies detect AI-generated images by exploiting low-level forensic cues left by the generation process. Frequency-based detectors exploit spectral artifacts introduced by upsampling~\cite{frank2020FreDect,karageorgiou2025Spec_SPAL} or learn representations directly in the frequency domain~\cite{tan2024frequencyFreqNet}. NPR~\cite{tan2024rethinkingNPR} instead examines correlations between neighboring pixels. DIRE~\cite{wang2023dire} uses the reconstruction error of a pretrained diffusion model as a detection cue, while DRCT~\cite{chen2024drct} uses diffusion reconstructions as hard examples for contrastive training. These traces can vary across generators and may be weakened by common post-processing operations such as compression and resizing~\cite{li2025improvingSAFE}.
Other studies build detectors using features from pretrained vision models. UnivFD~\cite{ojha2023towards} trains a linear classifier on frozen CLIP features~\cite{radford2021learningclip}. Zhou et al.~\cite{zhou2026simplicity} show that a linear classifier on frozen Perception Encoder (PE) features~\cite{bolya2026perceptionPEcore} can also generalize to unseen generators. Other methods adapt pretrained models to the detection task. FatFormer~\cite{liu2024fatformer} introduces forgery-aware adapters, Effort~\cite{yan2025effort} restricts adaptation to the subspace orthogonal to the principal components, and IAPL~\cite{li2026towardsIAPL} adjusts the encoder's prompts for each input image. These approaches may underuse the rich visual representations learned during pretraining, limiting generalization to unseen scenarios.

\begin{figure*}[!t]
  \centering
  \includegraphics[width=\textwidth]{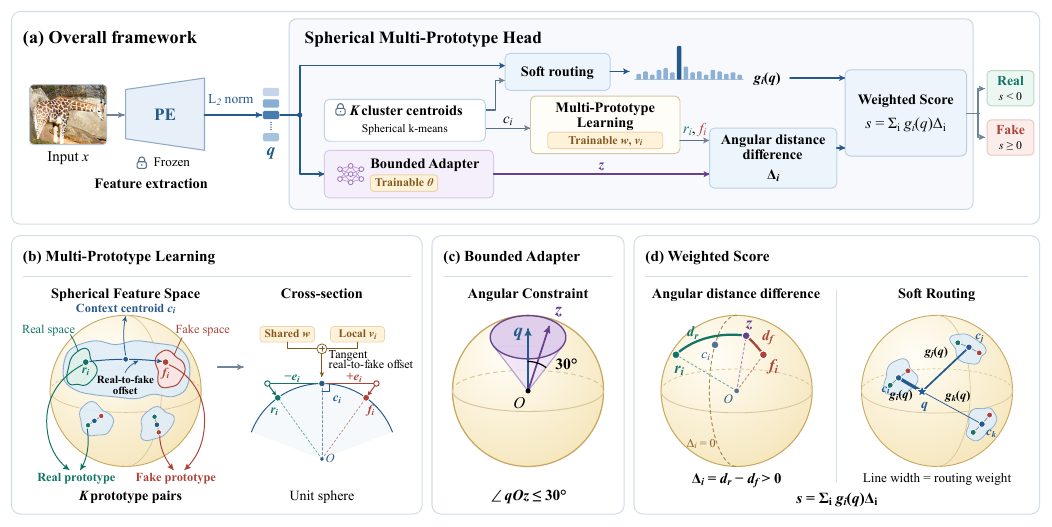}
\caption{\textbf{Overview of PE-SMP.}
\textbf{(a) Overall framework.} A frozen Perception Encoder extracts an
image feature, and the SMP head uses it to produce a detection score.
\textbf{(b) Multi-prototype learning.} The head learns a real and a fake
prototype for each visual context using a shared offset and a
context-specific offset.
\textbf{(c) Bounded adapter.} The adapter refines the image feature
while limiting its angular change on the hypersphere.
\textbf{(d) Weighted score.} The head compares the angular distances
from the adapted feature to the real and fake prototypes in each context.
It combines these differences, giving more weight to contexts whose
centroids are more similar to the original feature.}
  \label{fig:smp-overview}
\end{figure*}

\subsection{Prototype learning}

Prototype learning represents each class with one or more representative vectors and classifies samples according to their distances to these prototypes. Prototypical networks~\cite{snell2017protoNet} compute class prototypes as mean embeddings for few-shot classification, and convolutional prototype networks~\cite{yang2020convolutional_CPN} learn multiple prototypes per class to improve robustness in open-set recognition. When features are normalized, prototypes naturally lie on a hypersphere and similarity is measured by angles. Hyperspherical prototype networks~\cite{mettes2019hyperspherical_HPN} position class prototypes on the unit hypersphere with maximal separation, and normalized softmax and angular margin losses~\cite{wang2017normface,liu2017sphereface,wang2018cosface,deng2019arcface} treat classifier weights as hyperspherical prototypes and enlarge the angular margins between classes. In AI-generated image detection, MiraGe~\cite{shi2025mirage} uses text embeddings as class anchors to compact features within each class and separate real from fake images. However, a single global pair of real and fake prototypes provides the same reference points for all images, making it difficult to capture how the separation between real and generated features varies across visual contexts.

\section{Method}
\label{sec:method}

\subsection{Overview}
\label{sec:smp-readout}

PE-SMP pairs a frozen Perception Encoder with the SMP detection head
(\cref{fig:smp-overview}(a)). The encoder produces an image-level feature
that the head uses to assign a weight to each visual context. Each context
has a pair of real and fake prototypes that model the real-to-fake offset
around its cluster centroid. The head compares the adapted feature with
each prototype pair and combines the resulting angular distance differences
using the context weights to produce a single detection score.

\paragraph{Feature extraction.}
Bolya et al.~\cite{bolya2026perceptionPEcore} show that intermediate PE
features can outperform final embeddings on downstream tasks.
Motivated by this finding, we extract the patch tokens from block 50 of
PE-Core G/14, before the encoder's output normalization, attention
pooling, and visual projection.
For an input image $x$, we average its $N$ patch tokens
$t_j(x)\in\mathbf{R}^{1536}$, then project and normalize the pooled feature
to obtain
\begin{equation}
  h = Q^\top\!\left(\frac{1}{N}\sum_{j=1}^{N} t_j(x)\right),
  \qquad q = \operatorname{unit}(h),
  \label{eq:smp-readout}
\end{equation}
where $\operatorname{unit}(h)=h/\lVert h\rVert_2$ and the columns of
$Q\in\mathbf{R}^{1536\times1280}$ are orthonormal, obtained by QR
factorization of a Gaussian matrix. The encoder and $Q$ are fixed.
The resulting vector $q\in\mathbf{S}^{1280}$ feeds the hyperspherical head.
Here $\mathbf{S}^{1280}$ denotes the unit sphere in $\mathbf{R}^{1280}$.

\paragraph{Hyperspherical head.}
Each fixed cluster centroid $c_i$ defines a context and anchors a pair of
learned real and fake prototypes $(r_i,f_i)$. Soft routing uses the cosine
similarity between $q$ and each $c_i$ to compute the weights $g_i(q)$.
In parallel, the bounded adapter maps $q$ to a nearby unit feature $z$.
For each pair, the head subtracts the angular distance from $z$ to $f_i$
from the angular distance to $r_i$, giving $\Delta_i$. The routing-weighted
sum of these differences yields the detection score $s$. We classify the
image as fake when $s\geq 0$.

\subsection{Spherical multi-prototype learning}
\label{sec:smp-prototypes}

We model real-to-fake differences within high-level visual contexts in
the pretrained feature space. The offset analysis reveals a component
shared across classes and deviations that vary by class. Motivated by
this structure, we learn a shared offset across the discovered contexts
and a local offset for each context. See the supplementary section
\emph{Cross-class offset alignment} for details.

\paragraph{Cluster centroids.}
The pooled feature $q$ describes the image as a whole using patch tokens
from a late block of the pretrained encoder. Spherical
$k$-means~\cite{dhillon2001concept} groups these representations by cosine
similarity, forming contexts in the high-dimensional feature space
without predefined category labels. We cluster the $q$ features of 4,096
source training images jointly, with equal numbers of real and fake
images. We fix the resulting $K=16$ unit centroids $\{c_i\}_{i=1}^K$.
Each centroid provides a common reference for the real and fake
prototypes in its context.
\paragraph{Paired prototypes.}
The shared offset $w$ contributes to all prototype pairs, and the local
offset $v_i$ adjusts it for context $i$. As illustrated in
\cref{fig:smp-overview}(b), we project their sum onto the tangent space
at $c_i$ and rescale the result to obtain a bounded offset $e_i$.
\begin{equation}
  u_i = \Pi_{c_i}(w+v_i), \qquad
  e_i = \frac{u_i}{\sqrt{1+\lVert u_i\rVert_2^2}},
  \label{eq:smp-tangent}
\end{equation}
where $\Pi_{c_i}$ keeps the component of $w+v_i$ perpendicular to the
radius through $c_i$.
The rescaling keeps $\lVert e_i\rVert_2$ below one. We subtract $e_i$ from $c_i$ for the
real prototype and add it for the fake prototype. Normalizing the two
displaced vectors gives
\begin{equation}
  r_i=\operatorname{unit}(c_i-e_i), \qquad
  f_i=\operatorname{unit}(c_i+e_i).
  \label{eq:smp-pairs}
\end{equation}
We apply this construction at every cluster centroid, learning one real
and one fake prototype per context. The direction of $e_i$ determines
the orientation of the pair, and its norm controls their angular
separation. The prototypes lie symmetrically on either side of $c_i$,
each at an angular distance of $\arctan\lVert e_i\rVert_2$ from the
centroid. Thus $c_i$ is the midpoint of the spherical arc between them.

\subsection{Bounded feature adaptation}
\label{sec:smp-adapter}

The adapter refines $q$ for comparison with the prototypes. We bound the
rotation on the sphere to keep adapted features close to their original
positions (\cref{fig:smp-overview}(c)). This discourages a collapse
during training in which features across contexts converge toward a single
prototype pair and the detector relies too heavily on that pair.
A bias-free two-layer MLP $M_\theta$ with a GELU activation predicts a
residual, which we project onto the tangent space at $q$. With
$B=\tan(30^\circ)$, we set
\begin{equation}
  \begin{aligned}
    t &= \Pi_q\!\left(0.3\,M_\theta(q)\right), \\
    z &= \operatorname{unit}\!\left(q+
      \frac{B\,t}{\sqrt{B^2+\lVert t\rVert_2^2}}\right).
  \end{aligned}
  \label{eq:smp-residual}
\end{equation}
The added vector is orthogonal to $q$ and has norm at most $B$, which
bounds the angle $\angle qOz$ by $30^\circ$, where $O$ is the origin.
We initialize the MLP's last layer to zero, so the initial adapted
feature is $z=q$.

\subsection{Soft-routed angular score}
\label{sec:smp-decision}

\paragraph{Angular distance difference.}
On the unit sphere, the shortest arc length between two points equals
their central angle in radians. For unit vectors $a$ and $b$, the dot
product $a^\top b$ gives the cosine of this angle. Applying the inverse
cosine function $\arccos$ gives their spherical distance
$d_{\mathbf{S}}(a,b)=\arccos(a^\top b)$. Within context $i$, we measure
$d_r=d_{\mathbf{S}}(z,r_i)$ and $d_f=d_{\mathbf{S}}(z,f_i)$
(\cref{fig:smp-overview}(d)). The angle difference $\Delta_i=d_r-d_f$
thus gives the signed difference in spherical distances from $z$ to the
two prototypes. It is positive when $z$ is closer to the fake prototype
and zero when the distances are equal.

\paragraph{Soft routing and aggregation.}
In parallel, we compute the routing weights from the original feature
$q$ by applying a softmax to its scaled cosine similarities to the fixed
centroids.
\begin{equation}
  g_i(q)=\frac{\exp(\beta c_i^\top q)}
               {\sum_{j=1}^K\exp(\beta c_j^\top q)},
  \label{eq:smp-routing}
\end{equation}
where the scale $\beta$ controls how strongly the weights concentrate on
the most similar centroids. We fix $\beta=8$.
For a given image, the frozen encoder and fixed centroids determine its
routing weights throughout training. The adapter affects the score
through $z$, which is used to compute the angular distance differences
$\Delta_i$. We combine these differences across all contexts to obtain
\begin{equation}
  s(x)=\sum_{i=1}^K g_i(q)\,\Delta_i,
  \qquad \widehat y=\mathbf{1}\{s(x)\geq 0\},
  \label{eq:smp-score}
\end{equation}
with $\widehat y=0$ for real and $\widehat y=1$ for fake.

\subsection{Initialization and training}
\label{sec:smp-training}

\paragraph{Prototype initialization.}
We assign each training image used to fit the cluster centroids to its
nearest centroid $c_i$. Within cluster $i$, we average the unit features
$q$ of real and fake images separately to obtain two feature centroids.
We use the vector from the real feature centroid to the fake feature
centroid to initialize the prototype pair $(r_i,f_i)$. If either group is
absent from a cluster, we compute this vector using the real and fake
feature centroids over the full set of initialization images.

\paragraph{Training objective.}
We jointly train $\theta$, $w$, and $\{v_i\}_{i=1}^K$ using binary
labels $y$ on the source images, with $y=0$ for real and $y=1$ for fake.
We minimize the weighted sum
\begin{equation}
  \mathcal L = \mathcal L_{\mathrm{margin}}
    + 0.1\,\mathcal L_{\mathrm{cap}}
    + 0.01\,\mathcal L_{\mathrm{anchor}}.
  \label{eq:smp-loss}
\end{equation}
The margin loss targets $s(x)\leq -0.1$ for real images and
$s(x)\geq 0.1$ for fake images. The per-image penalty is
\begin{equation}
  \ell_{\mathrm{margin}}(x,y)=
  \begin{cases}
    \max\{0, s(x)+0.1\}, & y=0\ \text{(real)}, \\
    \max\{0, 0.1-s(x)\}, & y=1\ \text{(fake)}.
  \end{cases}
  \label{eq:smp-margin-loss}
\end{equation}
For the cap loss, $A(x,y)$ is the routing-weighted mean cosine similarity
between the adapted feature $z$ and the prototype matching label $y$ in
each context. The per-image penalty is
$\ell_{\mathrm{cap}}(x,y)=\max\{0, 0.5-A(x,y)\}$.
We average the per-image penalties over the training images to obtain
$\mathcal L_{\mathrm{margin}}$ and $\mathcal L_{\mathrm{cap}}$, respectively.

The anchor loss averages the squared Euclidean distances from each
real and fake prototypes to their respective fixed cluster centroids
$c_i$, penalizing prototype drift.

\section{Experiments}
\label{sec:experiments}

\subsection{Experimental setup}

\noindent
\textbf{Evaluation settings.}
To comprehensively evaluate the generalization ability of PE-SMP across different training sources, generative models, and application scenarios, we adopt three training and evaluation settings.

\textbf{GenImage setting.}
Following~\cite{zhu2023genimage,PGCzhou26h,zhou2026simplicity}, we train the detector solely on the SD v1.4 subset of GenImage and evaluate it on benchmarks spanning three scenarios.
(1) In the standard scenario, we evaluate on the 8 test subsets of GenImage, including the in-distribution SD v1.4 subset and 7 unseen subsets.
(2) For the in-the-wild scenario, we evaluate on Chameleon~\cite{yan2025sanityAIDE}, CommunityAI~\cite{CommunityAI}, SocialRF~\cite{wildRF}, and WildRF~\cite{wildRF}, whose images are collected from online creative communities and social media platforms.
(3) In the unseen-generator scenario, we evaluate on AIGI-Holmes~\cite{zhou2025holmes} and AIGI-Now~\cite{chen2025taskAIGInow}.
AIGI-Holmes covers 10 recent diffusion and autoregressive generators, with one test subset per generator.
AIGI-Now contains 18 test subsets from nine recent commercial and open-source generators, with one pixel-level and one semantic-level subset per generator.

\textbf{AIGCDetectBenchmark setting.}
Following~\cite{guo2025omniaid,zhong2023patchcraftAIGD}, we train the detector on the ProGAN training set covering 20 semantic categories and evaluate it on the 17 test subsets of AIGCDetectBenchmark~\cite{zhong2023patchcraftAIGD}.

\textbf{UniversalFakeDetect setting.}
Following~\cite{tan2024rethinkingNPR,li2026towardsIAPL}, we train the detector on ProGAN images from four semantic categories (car, cat, chair, and horse) and evaluate it on the 19 test subsets of UniversalFakeDetect~\cite{ojha2023towards}.

\noindent
\textbf{Metrics.}
Following prior work~\cite{PGCzhou26h,li2026towardsIAPL}, we report classification accuracy (Acc) across all three settings.
Acc is averaged across the real and AI-generated classes.
The supplementary material reports average precision (AP) for the GenImage and UniversalFakeDetect settings, treating AI-generated images as the positive class.

\noindent
\textbf{Implementation details.}
Across all three evaluation settings, we use the PE-SMP configuration described in \cref{sec:method}.
We train the SMP head using AdamW with a batch size of 512.
For the bounded adapter parameters, we use a learning rate of $3{\times}10^{-3}$ and a weight decay of 0.01.
For the shared and local prototype offsets, we use a learning rate of $3{\times}10^{-2}$ and zero weight decay.
Both learning rates follow cosine schedules.
All experiments are conducted on a single NVIDIA H200 GPU.

\subsection{Generalization comparisons}
\label{sec:exp-generalization}

\noindent
\textbf{Evaluation on the GenImage setting.}
\Cref{tab:cross_bench} reports Acc. See the supplementary section \emph{Additional benchmark results} for AP.
PE-SMP achieves the best accuracy on six of the seven benchmarks, with an average of 97.19\% that surpasses the second-best method by 23.88\%, and its accuracy never falls below 95\%.
In the standard scenario, most recent detectors perform well, and PE-SMP reaches 98.62\%, second only to PGC~\cite{PGCzhou26h} at 99.62\%.
However, this nearly saturated benchmark poorly reflects generalization.
On the in-the-wild benchmarks, test images differ from the training data in both content and post-processing.
Most baselines achieve accuracies of only 46\% to 75\%.
PE-SMP attains between 95.03\% and 98.29\% and outperforms the strongest baseline on each benchmark by 13.83\% to 27.25\%.
This gap may be attributed to the fact that post-processing weakens the low-level traces used by existing detectors.
In the unseen-generator scenario, PE-SMP achieves 99.65\% accuracy on AIGI-Holmes and 95.98\% on AIGI-Now.
These results indicate that PE-SMP generalizes well from a single training generator to a diverse set of unseen generators, including diffusion and autoregressive models.

\noindent
\textbf{Evaluation on the AIGCDetectBenchmark setting.}
The Acc results are shown in \cref{tab:aigcdetect}.
PE-SMP achieves the best mAcc of 96.62\%, exceeding AIDE by 3.60\%, and ranks first or second on 11 of the 17 subsets.
Effort and OmniAID exceed 92\% on every GAN subset but average only about 60\% on the diffusion subsets, whereas PE-SMP averages 99.18\% and 94.34\% on the two groups, respectively.
PE-SMP is less competitive on Midjourney and ADM, where SAFE and AIDE perform best.

\noindent
\textbf{Evaluation on the UniversalFakeDetect setting.}
\Cref{tab:ufd} reports Acc, with AP in the same supplementary section.
PE-SMP achieves 94.65\% mAcc and 98.86\% mAP, second only to IAPL  at 95.61\% and 99.32\%.
The gap mainly comes from the low-level vision subsets SITD and SAN.
Their fake images are restored from real inputs and share the same semantic content, which leaves mostly pixel-level differences.

\begin{table*}[t]
  \centering
  \caption{\textbf{Cross-benchmark evaluation on seven detection benchmarks using Accuracy (\%).}
    $^\clubsuit$: retrained using the official code under the same training setting.
    $^\spadesuit$: evaluated using the publicly released weights.
    \textbf{Following the GenImage protocol, all models are trained solely on the SDv1.4 subset and then evaluated on each benchmark to assess cross-dataset generalization.}
    The best and second-best results are in \best{bold} and \second{underlined}, respectively.}
  \label{tab:cross_bench}
  \setlength{\tabcolsep}{6pt}
  \resizebox{\textwidth}{!}{%
  \begin{tabular}{lc|ccccccc|c}
    \toprule
    Methods & Venues & GenImage & Chameleon & CommunityAI & SocialRF & WildRF & AIGI-Holmes & AIGI-Now & Avg. \\
    \midrule
    NPR$^\spadesuit$~\cite{tan2024rethinkingNPR} & CVPR 2024 & 88.27 & 61.24 & 59.32 & 59.40 & 63.25 & 86.79 & 70.29 & 69.79 \\
    DRCT$^\spadesuit$~\cite{chen2024drct} & ICML 2024 & 82.08 & 59.75 & 67.05 & \second{67.78} & \second{74.03} & 75.95 & 58.56 & 69.32 \\
    OMAT$^\spadesuit$~\cite{zhou2026breakingOMAT} & NeurIPS 2025 & 94.54 & 62.86 & 63.33 & 58.08 & 67.22 & 87.72 & 61.51 & 70.75 \\
    SAFE$^\clubsuit$~\cite{li2025improvingSAFE} & KDD 2025 & 84.31 & 51.13 & 53.12 & 56.72 & 59.00 & 75.46 & 63.58 & 63.33 \\
    AIDE$^\spadesuit$~\cite{yan2025sanityAIDE} & ICLR 2025 & 87.14 & 59.36 & 61.93 & 60.48 & 63.61 & 80.55 & 66.23 & 68.47 \\
    Effort$^\spadesuit$~\cite{yan2025effort} & ICML 2025 & 90.11 & 58.55 & 59.83 & 61.03 & 71.46 & 83.33 & 63.31 & 69.66 \\
    C2P-CLIP$^\spadesuit$~\cite{tan2025c2p} & AAAI 2025 & 96.03 & 51.29 & 46.69 & 46.45 & 48.93 & \second{92.99} & \second{82.74} & 66.45 \\
    VIB-Net$^\spadesuit$~\cite{zhang2025towardsVIB} & CVPR 2025 & 84.79 & 61.85 & 60.36 & 55.73 & 64.03 & 86.16 & 76.49 & 69.92 \\
    OmniAID$^\spadesuit$~\cite{guo2025omniaid} & ICML 2026 & 94.03 & \second{79.20} & \second{82.83} & 57.32 & 59.88 & 80.92 & 59.00 & \second{73.31} \\
    PGC$^\spadesuit$~\cite{PGCzhou26h} & ICML 2026 & \best{99.62} & 53.44 & 55.82 & 60.60 & 62.51 & 85.14 & 66.51 & 69.09 \\
    IAPL$^\spadesuit$~\cite{li2026towardsIAPL} & CVPR 2026 & 96.78 & 76.63 & 75.87 & 50.63 & 55.11 & 64.89 & 59.94 & 68.55 \\
    \midrule
    \rowcolor{oursrow}
    PE-SMP (Ours) & -- & \second{98.62} & \best{96.07} & \best{96.66} & \best{95.03} & \best{98.29} & \best{99.65} & \best{95.98} & \best{97.19} \\
    \bottomrule
  \end{tabular}%
  }
\end{table*}

\begin{table*}[t]
  \centering
  \caption{\textbf{Evaluation on the AIGCDetectBenchmark~\cite{zhong2023patchcraftAIGD} dataset using Accuracy (\%).}
    $^\clubsuit$: retrained using the official code under the same training setting.
    The results of NPR and AIDE are cited from AIDE~\cite{yan2025sanityAIDE}, and those of Effort are cited from OmniAID~\cite{guo2025omniaid}.
    \textbf{The ProGAN is used as the training data.}
    The best and second-best results are in \best{bold} and \second{underlined}, respectively.}
  \label{tab:aigcdetect}
  \setlength{\tabcolsep}{2.2pt}
  \resizebox{\textwidth}{!}{%
  \begin{tabular}{l|cccccccc|ccccccccc|c}
    \toprule
    \multirow{2}{*}{Methods}
      & \multicolumn{8}{c}{GAN-based}
      & \multicolumn{9}{|c|}{Diffusion-based}
      & \multirow{2}{*}{mAcc} \\
    \cmidrule(lr){2-9} \cmidrule(lr){10-18}
      & \tworow{Pro-}{GAN} & \tworow{Style-}{GAN} & \tworow{Big-}{GAN} & \tworow{Cycle-}{GAN} & \tworow{Star-}{GAN} & \tworow{Gau-}{GAN} & \tworow{Style-}{GAN2} & WFIR
      & ADM & GLIDE & \tworow{Mid-}{journey} & \tworow{SD}{v1.4} & \tworow{SD}{v1.5} & VQDM & Wukong & \tworow{DALLE}{2} & SDXL & \\
    \midrule
    FreqNet$^\clubsuit$~\cite{tan2024frequencyFreqNet}
      & 89.08 & 66.97 & 75.20 & 47.62 & 48.37 & 56.61 & 69.49 & 46.80
      & 51.79 & 36.79 & 43.99 & 46.84 & 46.64 & 53.91 & 48.18 & 31.70 & 58.43 & 54.02 \\
    NPR~\cite{tan2024rethinkingNPR}
      & 99.79 & 97.70 & 84.35 & 96.10 & 99.35 & 82.50 & \second{98.38} & 65.80
      & 69.69 & 78.36 & \second{77.85} & 78.63 & 78.89 & 78.13 & 76.11 & 64.90 & 94.10 & 83.57 \\
    RINE$^\clubsuit$~\cite{koutlis2024RINE}
      & 99.99 & 96.53 & 99.43 & \second{99.58} & 99.15 & 99.74 & 92.25 & 97.70
      & 73.51 & 75.48 & 57.20 & 87.40 & 86.98 & 88.21 & 87.43 & 55.25 & 65.45 & 85.96 \\
    AIDE~\cite{yan2025sanityAIDE}
      & 99.99 & \best{99.64} & 83.95 & 98.48 & \second{99.91} & 73.25 & 98.00 & 94.20
      & \best{93.43} & \second{95.09} & 77.20 & 93.00 & 92.85 & \second{95.16} & 93.55 & \second{96.60} & \second{97.05} & \second{93.02} \\
    C2P-CLIP$^\clubsuit$~\cite{tan2025c2p}
      & 100.00 & 96.61 & 99.10 & 99.17 & 99.87 & 99.09 & 94.62 & 77.20
      & 69.06 & 88.87 & 57.95 & 92.63 & 92.41 & 88.58 & 89.69 & 67.05 & 79.70 & 87.74 \\
    SAFE$^\clubsuit$~\cite{li2025improvingSAFE}
      & 97.24 & 96.54 & 80.63 & 93.87 & 97.42 & 77.16 & 94.89 & 52.45
      & 84.51 & \best{95.61} & \best{94.45} & \second{95.67} & \second{95.35} & 90.92 & 93.02 & \best{97.25} & 96.60 & 90.21 \\
    Effort~\cite{yan2025effort}
      & \second{100.00} & 95.80 & \second{99.58} & \best{99.66} & \best{99.98} & 99.84 & 92.55 & 94.60
      & 70.68 & 64.61 & 50.03 & 55.23 & 55.21 & 76.55 & 56.77 & 53.05 & 50.13 & 77.31 \\
    OmniAID$^\clubsuit$~\cite{guo2025omniaid}
      & 100.00 & 98.71 & 99.58 & 99.77 & 99.85 & 99.92 & 92.88 & 94.70
      & 73.46 & 64.61 & 50.38 & 55.22 & 55.28 & 80.28 & 58.18 & 55.70 & 51.50 & 78.24 \\
    IAPL$^\clubsuit$~\cite{li2026towardsIAPL}
      & 99.99 & 96.31 & 99.28 & 98.33 & 99.05 & 98.48 & 97.53 & \second{99.35}
      & 80.58 & 92.32 & 56.50 & 93.19 & 92.79 & 90.07 & \second{93.87} & 75.35 & 77.70 & 90.63 \\
    PGC$^\clubsuit$~\cite{PGCzhou26h}
      & \best{100.00} & 85.05 & 99.30 & 98.83 & 63.36 & \second{99.93} & 89.42 & 67.65
      & 66.40 & 68.96 & 51.43 & 71.46 & 70.36 & 88.24 & 75.46 & 55.65 & 72.75 & 77.90 \\
    \midrule
    \rowcolor{oursrow}
    PE-SMP (Ours)
      & 99.93 & \second{98.71} & \best{99.65} & 97.99 & 98.65 & \best{99.95} & \best{99.07} & \best{99.50}
      & \second{86.73} & 94.50 & 77.14 & \best{98.58} & \best{98.41} & \best{99.10} & \best{99.33} & 96.40 & \best{98.85} & \best{96.62} \\
    \bottomrule
  \end{tabular}%
  }
\end{table*}

\begin{table*}[t]
  \centering
  \caption{\textbf{Evaluation on the UniversalFakeDetect dataset~\cite{ojha2023towards} using Accuracy (\%).}
    Results of the other methods are cited from IAPL~\cite{li2026towardsIAPL}.
    $^\clubsuit$: retrained using the official code under the same training setting.
    \textbf{The 4-class ProGAN is used as the training data.}
    The best and second-best results are in \best{bold} and \second{underlined}, respectively.}
  \label{tab:ufd}
  \setlength{\tabcolsep}{2.2pt}
  \resizebox{\textwidth}{!}{%
  \begin{tabular}{l|ccccccccccccccccccc|c}
    \toprule
    \multirow{2}{*}{Methods}
      & \multicolumn{6}{c}{GAN}
      & \multirow{2}{*}{\tworow{Deep}{fakes}}
      & \multicolumn{2}{c}{Low level}
      & \multicolumn{2}{c}{Perceptual loss}
      & \multirow{2}{*}{Guided}
      & \multicolumn{3}{c}{LDM}
      & \multicolumn{3}{c}{Glide}
      & \multirow{2}{*}{Dalle}
      & \multirow{2}{*}{mAcc} \\
    \cmidrule(lr){2-7} \cmidrule(lr){9-10} \cmidrule(lr){11-12} \cmidrule(lr){14-16} \cmidrule(lr){17-19}
      & \tworow{Pro-}{GAN} & \tworow{Cycle-}{GAN} & \tworow{Big-}{GAN}
      & \tworow{Style-}{GAN} & \tworow{Gau-}{GAN} & \tworow{Star-}{GAN}
      &
      & SITD & SAN
      & CRN & IMLE
      &
      & \tworow{200}{steps} & \tworow{200}{w/cfg} & \tworow{100}{steps}
      & \tworow{100}{27} & \tworow{50}{27} & \tworow{100}{10}
      & & \\
    \midrule
    FreqNet~\cite{tan2024frequencyFreqNet}
      & 97.90 & 95.84 & 90.45 & 97.55 & 90.24 & 93.41
      & \best{97.40} & 88.92 & 59.04 & 71.92 & 67.35 & 86.70
      & 84.55 & \best{99.58} & 65.56 & 85.69 & 97.40 & 88.15 & 59.06 & 85.09 \\
    NPR~\cite{tan2024rethinkingNPR}
      & 99.84 & 95.00 & 87.55 & 96.23 & 86.57 & 99.75
      & 76.89 & 66.94 & \best{98.63} & 50.00 & 50.00 & 84.55
      & 97.65 & 98.00 & 98.20 & 96.25 & 97.15 & 97.35 & 87.15 & 87.56 \\
    RINE~\cite{koutlis2024RINE}
      & 100.00 & 99.30 & \best{99.60} & 88.90 & 99.80 & 99.50
      & 80.60 & 90.60 & 68.30 & 89.20 & 90.60 & 76.10
      & 98.30 & 88.20 & 98.60 & 88.90 & 92.60 & 90.70 & 95.00 & 91.31 \\
    AIDE~\cite{yan2025sanityAIDE}
      & 99.99 & 98.49 & 83.95 & \best{99.65} & 73.24 & 99.90
      & 54.09 & 68.06 & 71.14 & 60.87 & 61.00 & \best{88.50}
      & 98.20 & 97.45 & 98.40 & \second{98.15} & \best{98.40} & 97.85 & 97.50 & 86.57 \\
    C2P-CLIP~\cite{tan2025c2p}
      & 99.98 & 97.31 & 99.12 & 96.44 & 99.17 & 99.60
      & 93.77 & \best{95.56} & 64.38 & \second{93.29} & \second{93.29} & 69.10
      & \second{99.25} & 97.25 & \second{99.30} & 95.25 & 95.25 & 96.10 & 98.55 & 93.79 \\
    MiraGe~\cite{shi2025mirage}
      & \second{100.00} & 94.30 & 96.50 & 96.80 & 93.60 & 96.10
      & 88.70 & 75.80 & 71.90 & 92.90 & 92.90 & 82.00
      & 98.30 & 94.60 & 98.60 & 97.50 & 97.50 & 98.00 & 98.60 & 92.87 \\
    SAFE~\cite{li2025improvingSAFE}
      & 99.86 & 98.87 & 89.73 & \second{98.04} & 91.52 & \second{99.90}
      & 93.10 & 85.56 & \second{95.91} & 50.10 & 50.10 & 82.35
      & 98.80 & \second{98.65} & 98.80 & 95.75 & 96.60 & 97.25 & 97.50 & 90.44 \\
    Effort$^\clubsuit$~\cite{yan2025effort}
      & 99.94 & \best{100.00} & 98.90 & 87.34 & \second{99.92} & \best{100.00}
      & 62.48 & 58.61 & 57.08 & 69.94 & 88.37 & 55.60
      & 94.40 & 78.10 & 94.80 & 72.35 & 71.95 & 71.75 & 90.60 & 81.69 \\
    PGC~\cite{PGCzhou26h}
      & 97.30 & 99.10 & 96.70 & 77.20 & 98.60 & 90.70
      & 77.20 & 54.40 & 91.10 & 88.10 & 88.40 & \second{87.50}
      & 92.70 & 98.10 & 97.60 & 97.00 & 96.70 & 97.00 & 97.00 & 90.65 \\
    IAPL~\cite{li2026towardsIAPL}
      & \best{100.00} & 98.60 & 98.65 & 94.89 & 99.39 & 96.70
      & \second{95.89} & \second{90.83} & 93.84 & 92.47 & 92.72 & 72.75
      & \best{99.50} & 97.70 & 99.15 & 97.95 & \second{98.30} & \best{98.35} & \best{98.90} & \best{95.61} \\
    \midrule
    \rowcolor{oursrow}
    PE-SMP (Ours)
      & 99.89 & 96.10 & 98.85 & 97.38 & \best{99.92} & 98.50
      & 93.06 & 79.17 & 67.12 & \best{99.59} & \best{99.76} & 85.15
      & 99.20 & 94.15 & \best{99.40} & \best{98.40} & 97.90 & \second{98.25} & 96.60 & \second{94.65} \\
    \bottomrule
  \end{tabular}%
  }
\end{table*}

\subsection{Robustness evaluation}

\label{sec:exp-robustness}

\begin{figure}[!b]
  \centering
  \includegraphics[width=\linewidth]{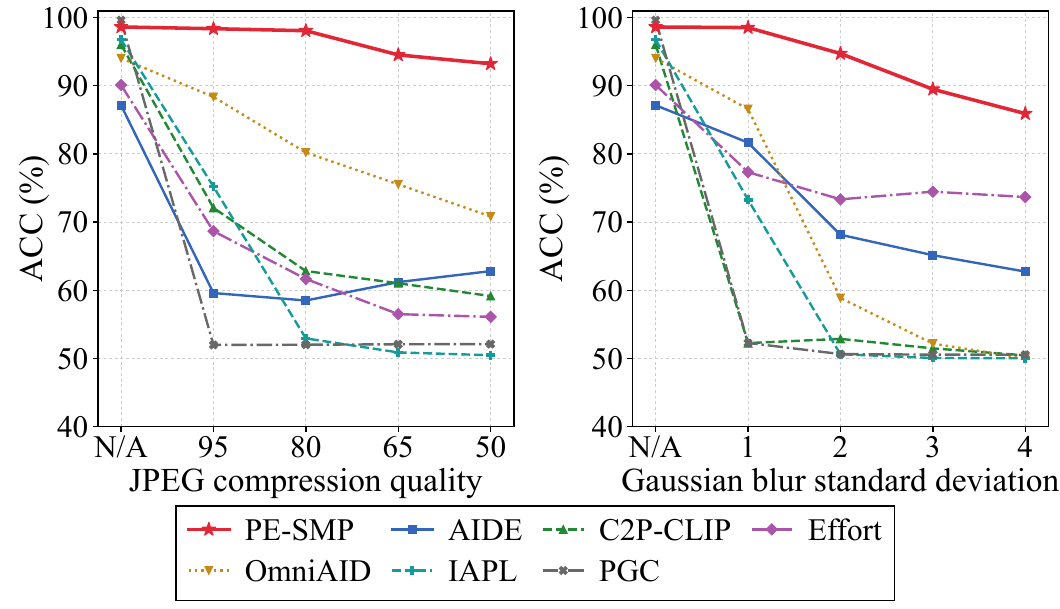}
  \caption{\textbf{Detection accuracy under JPEG compression and Gaussian blur on GenImage.}
    Both operations are applied only at test time.
    N/A denotes unperturbed images.}
  \label{fig:robustness}
\end{figure}

All seven detectors are trained on the GenImage SD v1.4 subset. We compare their accuracy on GenImage test images subjected to JPEG compression or Gaussian blur (\cref{fig:robustness}).
PE-SMP is trained without data augmentation and achieves the highest accuracy at every tested level of both perturbations.

At JPEG quality 50, PE-SMP achieves 93.2\% accuracy, 22.4 percentage points above OmniAID, the strongest baseline at that level.
At a Gaussian blur standard deviation of 4, PE-SMP achieves 85.9\% accuracy, 12.2 percentage points above Effort, the strongest baseline at that level.
See supplementary \emph{Additional robustness results} for all four perturbations.

\subsection{Ablation studies}

Unless otherwise specified, we train on GenImage SD v1.4 and average the ablation results over all seven benchmarks in the GenImage setting.

\noindent
\textbf{Ablation of PE-SMP.}
\Cref{tab:ablation_components_main} isolates the feature readout and three head components in \cref{fig:smp-overview}.
Using the native PE-G output reduces mAcc by 15.99 percentage points, with fake-image accuracy falling from 97.95\% to 69.57\%.
Freezing the prototype pairs at initialization reduces mAcc by 2.19 points.
Measuring angular distances directly from $q$ without the bounded adapter reduces mAcc by 2.42 points.
Using hard routing reduces mAcc by 3.77 points and mAP by 7.27 points.
See supplementary \emph{Additional ablations} for per-benchmark results.

\begin{table}[t]
  \centering
  \caption{Ablation of PE-SMP. Metrics are averaged over seven benchmarks. RAcc and FAcc denote accuracy on real and fake images, respectively.}
  \label{tab:ablation_components_main}
  \resizebox{\linewidth}{!}{%
  \tabcolsep=14 pt
  \begin{tabular}{l|cccc}
    \toprule
    Method & mAcc & mAP & RAcc & FAcc \\
    \midrule
    w/o res. MLP & 94.77 & 99.59 & 91.81 & 97.73 \\
    w/o proto. training & 95.00 & 99.54 & 92.48 & 97.51 \\
    w/o soft routing & 93.42 & 92.44 & 89.91 & 96.93 \\
    w/o block 50 & 81.20 & 96.02 & 92.83 & 69.57 \\
    \midrule
    \rowcolor{oursrow}
    PE-SMP (Ours) & \textbf{97.19} & \textbf{99.71} & \textbf{96.42} & \textbf{97.95} \\
    \bottomrule
  \end{tabular}
  }
\end{table}

\noindent
\textbf{Discriminative head.}
With the same block-50 features, SMP and the MLP head reach similar mAP, but SMP misclassifies fewer real images as fake (96.42\% vs.\ 77.14\% real-image accuracy), with a small drop in fake-image accuracy (\cref{tab:ablation_head_main}).

\begin{table}[t]
  \centering
  \caption{Head comparison using the same block-50 features. Metrics are averaged over seven benchmarks. RAcc and FAcc denote accuracy on real and fake images, respectively.}
  \label{tab:ablation_head_main}
  \resizebox{\linewidth}{!}{%
  \tabcolsep=13 pt
  \begin{tabular}{l|cccc}
    \toprule
    Method & mAcc & mAP & RAcc & FAcc \\
    \midrule
     w/ Linear head & 88.08 & 98.98 & 77.35 & 98.80 \\
     w/ MLP head & 88.25 & 99.66 & 77.14 & \textbf{99.37} \\
    \midrule
\rowcolor{oursrow}
    PE-SMP (Ours) & \textbf{97.19} & \textbf{99.71} & \textbf{96.42} & 97.95 \\
    \bottomrule
  \end{tabular}
  }
\end{table}

\noindent
\textbf{Head comparison across frozen backbones.}
With input features fixed within each backbone configuration, replacing the linear head with SMP improves mean accuracy across the seven benchmarks in all five configurations of CLIP, DINOv3, and PE, with gains ranging from 4.17 to 25.63 percentage points (\cref{tab:ablation_backbone_scale_main}).
On PE-G, the 9.11-point gain comes primarily from fewer real images being classified as fake.
PE-SMP achieves the highest mean accuracy among the tested configurations (97.19\%).


\begin{table}[t]
  \centering
  \caption{Backbone comparison with a linear head and the SMP head on the same input. Metrics are averaged over seven benchmarks. RAcc and FAcc denote accuracy on real and fake images, respectively. Subscripts give the change of the SMP head over the linear head on the same backbone. The PE-G configuration with the SMP head is PE-SMP.}
  \label{tab:ablation_backbone_scale_main}
  \resizebox{\linewidth}{!}{%
  \tabcolsep=6pt
  \begin{tabular}{l|ll|cccc}
    \toprule
    Backbone & Scale & Head & mAcc & mAP & RAcc & FAcc \\
    \midrule
    CLIP & L & Linear & 67.58 & 72.82 & 73.88 & 61.28 \\
    CLIP & L & SMP & 71.75\up{4.17} & 77.81\up{4.99} & 78.33 & 65.17 \\
    \cmidrule(lr){1-7}
    DINOv3 & L & Linear & 70.01 & 93.50 & 46.65 & 93.36 \\
    DINOv3 & L & SMP & 85.01\up{15.00} & 94.47\up{0.97} & 85.58 & 84.44 \\
    \cmidrule(lr){1-7}
    PE & L & Linear & 86.37 & 94.99 & 95.15 & 77.60 \\
    PE & L & SMP & 92.18\up{5.81} & 98.68\up{3.69} & 94.41 & 89.95 \\
    \cmidrule(lr){1-7}
    DINOv3 & H+ & Linear & 63.87 & 94.47 & 32.33 & 95.42 \\
    DINOv3 & H+ & SMP & 89.50\up{25.63} & 96.24\up{1.77} & 91.07 & 87.92 \\
    \midrule
    PE & G & Linear & 88.08 & 98.98 & 77.35 & \textbf{98.80} \\
    \rowcolor{oursrow}
    PE & G & SMP & \textbf{97.19}\up{9.11} & \textbf{99.71}\up{0.73} & \textbf{96.42} & 97.95 \\
    \bottomrule
  \end{tabular}
  }
\end{table}

\noindent
\textbf{Hyperparameters.}
In \cref{tab:abl_angle,tab:abl_num_contexts}, a $30^\circ$ cap and $K{=}16$ yield the highest Acc and AP on Chameleon in their respective sweeps.
Acc varies more than AP.
The supplement reports full sweeps and class-wise accuracy.


\begin{table}[t]
  \centering
  \begin{minipage}[t]{0.49\linewidth}
    \centering
    \caption{\textbf{Residual angle cap on Chameleon.}}
    \label{tab:abl_angle}
    \footnotesize
    \setlength{\tabcolsep}{6pt}
    \begin{tabular}{c|cc}
      \toprule
      Angle & Acc & AP \\
      \midrule
      $10^\circ$ & 93.65 & 99.39 \\
      $20^\circ$ & 88.00 & 99.38 \\
      \rowcolor{oursrow}
      $30^\circ$ & \textbf{96.07} & \textbf{99.51} \\
      $45^\circ$ & 85.28 & 99.29 \\
      \bottomrule
    \end{tabular}
  \end{minipage}%
  \hfill
  \begin{minipage}[t]{0.49\linewidth}
    \centering
    \caption{\textbf{Number of contexts on Chameleon.}}
    \label{tab:abl_num_contexts}
    \footnotesize
    \setlength{\tabcolsep}{6pt}
    \begin{tabular}{c|cc}
      \toprule
      $K$ & Acc & AP \\
      \midrule
      8  & 95.82 & 98.99 \\
      \rowcolor{oursrow}
      16 & \textbf{96.07} & \textbf{99.51} \\
      32 & 90.69 & 99.39 \\
      64 & 73.87 & 99.18 \\
      \bottomrule
    \end{tabular}
  \end{minipage}
\end{table}

\section{Conclusion}

We found that class-wise real-to-fake offsets in PE-Core features exhibit a shared direction and class-specific residuals associated with image content. SMP models this structure with paired real and fake prototypes across visual contexts. Trained solely on the SD v1.4 subset of GenImage, PE-SMP generalizes to unseen generators and in-the-wild images while remaining robust to common post-processing.
\par

{\small
\bibliographystyle{ieeenat_fullname}
\bibliography{main}

@String(CVPR= {IEEE Conf. Comput. Vis. Pattern Recog.})

@String(ICCV= {Int. Conf. Comput. Vis.})

@String(AAAI = {AAAI})

@String(CVPR  = {CVPR})

@String(ICCV  = {ICCV})

@article{karras2017progressiveprogan,
  title={Progressive growing of gans for improved quality, stability, and variation},
  author={Karras, Tero and Aila, Timo and Laine, Samuli and Lehtinen, Jaakko},
  journal={arXiv preprint arXiv:1710.10196},
  year={2017}
}

@inproceedings{karras2019stylegan,
  title={A style-based generator architecture for generative adversarial networks},
  author={Karras, Tero and Laine, Samuli and Aila, Timo},
  booktitle={2019 IEEE/CVF conference on computer vision and pattern recognition (CVPR)},
  pages={4396--4405},
  year={2019},
  organization={IEEE}
}

@article{rombach2022highLDM,
  title={High-Resolution Image Synthesis with Latent Diffusion Models},
  author={Rombach, Robin},
  journal={arXiv preprint arXiv:2112.10752},
  year={2022}
}

@article{ramesh2022DALL-E3,
  title={Hierarchical text-conditional image generation with clip latents},
  author={Ramesh, Aditya and Dhariwal, Prafulla and Nichol, Alex and Chu, Casey and Chen, Mark},
  journal={arXiv preprint arXiv:2204.06125},
  volume={1},
  number={2},
  pages={3},
  year={2022}
}

@inproceedings{han2025infinity,
  title={{Infinity}$\infty$: Scaling bitwise autoregressive modeling for high-resolution image synthesis},
  author={Han, Jian and Liu, Jinlai and Jiang, Yi and Yan, Bin and Zhang, Yuqi and Yuan, Zehuan and Peng, Bingyue and Liu, Xiaobing},
  booktitle={2025 IEEE/CVF Conference on Computer Vision and Pattern Recognition (CVPR)},
  pages={15733--15744},
  year={2025},
  organization={IEEE}
}

@inproceedings{wu2025janus,
  title={Janus: Decoupling visual encoding for unified multimodal understanding and generation},
  author={Wu, Chengyue and Chen, Xiaokang and Wu, Zhiyu and Ma, Yiyang and Liu, Xingchao and Pan, Zizheng and Liu, Wen and Xie, Zhenda and Yu, Xingkai and Ruan, Chong and others},
  booktitle={2025 IEEE/CVF Conference on Computer Vision and Pattern Recognition (CVPR)},
  pages={12966--12977},
  year={2025},
  organization={IEEE}
}

@inproceedings{wang2023dire,
  title={Dire for diffusion-generated image detection},
  author={Wang, Zhendong and Bao, Jianmin and Zhou, Wengang and Wang, Weilun and Hu, Hezhen and Chen, Hong and Li, Houqiang},
  booktitle={2023 IEEE/CVF International Conference on Computer Vision (ICCV)},
  pages={22388--22398},
  year={2023},
  organization={IEEE}
}

@inproceedings{radford2021learningclip,
  title={Learning transferable visual models from natural language supervision},
  author={Radford, Alec and Kim, Jong Wook and Hallacy, Chris and Ramesh, Aditya and Goh, Gabriel and Agarwal, Sandhini and Sastry, Girish and Askell, Amanda and Mishkin, Pamela and Clark, Jack and others},
  booktitle={International conference on machine learning},
  pages={8748--8763},
  year={2021},
  organization={PmLR}
}

@inproceedings{karageorgiou2025Spec_SPAL,
  title={Any-resolution ai-generated image detection by spectral learning},
  author={Karageorgiou, Dimitrios and Papadopoulos, Symeon and Kompatsiaris, Ioannis and Gavves, Efstratios},
  booktitle={2025 IEEE/CVF Conference on Computer Vision and Pattern Recognition (CVPR)},
  pages={18706--18717},
  year={2025},
  organization={IEEE}
}

@inproceedings{frank2020FreDect,
  title={Leveraging frequency analysis for deep fake image recognition},
  author={Frank, Joel and Eisenhofer, Thorsten and Sch{\"o}nherr, Lea and Fischer, Asja and Kolossa, Dorothea and Holz, Thorsten},
  booktitle={International conference on machine learning},
  pages={3247--3258},
  year={2020},
  organization={PMLR}
}

@article{bolya2026perceptionPEcore,
  title={Perception encoder: The best visual embeddings are not at the output of the network},
  author={Bolya, Daniel and Huang, Po-Yao and Sun, Peize and Cho, Jang Hyun and Madotto, Andrea and Wei, Chen and Ma, Tengyu and Zhi, Jiale and Rajasegaran, Jathushan and Bangalath, Hanoona and others},
  journal={Advances in Neural Information Processing Systems},
  volume={38},
  pages={60884--60937},
  year={2026}
}

@inproceedings{snell2017protoNet,
  title={Prototypical networks for few-shot learning},
  author={Snell, Jake and Swersky, Kevin and Zemel, Richard},
  booktitle={Advances in neural information processing systems},
  volume={30},
  year={2017}
}

@article{yang2020convolutional_CPN,
  title={Convolutional prototype network for open set recognition},
  author={Yang, Hong-Ming and Zhang, Xu-Yao and Yin, Fei and Yang, Qing and Liu, Cheng-Lin},
  journal={IEEE Transactions on Pattern Analysis and Machine Intelligence},
  volume={44},
  number={5},
  pages={2358--2370},
  year={2020},
  publisher={IEEE}
}

@inproceedings{mettes2019hyperspherical_HPN,
  title={Hyperspherical prototype networks},
  author={Mettes, Pascal and Van der Pol, Elise and Snoek, Cees},
  booktitle={Advances in neural information processing systems},
  volume={32},
  year={2019}
}

@article{dhillon2001concept,
  title={Concept decompositions for large sparse text data using clustering},
  author={Dhillon, Inderjit S. and Modha, Dharmendra S.},
  journal={Machine Learning},
  volume={42},
  number={1},
  pages={143--175},
  year={2001},
  publisher={Springer}
}

@inproceedings{wang2017normface,
  title={Normface: L2 hypersphere embedding for face verification},
  author={Wang, Feng and Xiang, Xiang and Cheng, Jian and Yuille, Alan Loddon},
  booktitle={Proceedings of the 25th ACM international conference on Multimedia},
  pages={1041--1049},
  year={2017}
}

@inproceedings{liu2017sphereface,
  title={Sphereface: Deep hypersphere embedding for face recognition},
  author={Liu, Weiyang and Wen, Yandong and Yu, Zhiding and Li, Ming and Raj, Bhiksha and Song, Le},
  booktitle={2017 IEEE conference on computer vision and pattern recognition (CVPR)},
  pages={6738--6746},
  year={2017},
  organization={IEEE}
}

@inproceedings{wang2018cosface,
  title={Cosface: Large margin cosine loss for deep face recognition},
  author={Wang, Hao and Wang, Yitong and Zhou, Zheng and Ji, Xing and Gong, Dihong and Zhou, Jingchao and Li, Zhifeng and Liu, Wei},
  booktitle={2018 IEEE/CVF conference on computer vision and pattern recognition},
  pages={5265--5274},
  year={2018},
  organization={IEEE}
}

@inproceedings{deng2019arcface,
  title={Arcface: Additive angular margin loss for deep face recognition},
  author={Deng, Jiankang and Guo, Jia and Xue, Niannan and Zafeiriou, Stefanos},
  booktitle={Proceedings of the IEEE/CVF conference on computer vision and pattern recognition},
  pages={4690--4699},
  year={2019}
}

@inproceedings{shi2025mirage,
  title={MiraGe: Multimodal discriminative representation learning for generalizable AI-generated image detection},
  author={Shi, Kuo and Lu, Jie and Ye, Shanshan and Zhang, Guangquan and Fang, Zhen},
  booktitle={Proceedings of the 33rd ACM International Conference on Multimedia},
  pages={353--361},
  year={2025}
}

@inproceedings{ojha2023towards,
  title={Towards universal fake image detectors that generalize across generative models},
  author={Ojha, Utkarsh and Li, Yuheng and Lee, Yong Jae},
  booktitle={2023 IEEE/CVF Conference on Computer Vision and Pattern Recognition (CVPR)},
  pages={24480--24489},
  year={2023},
  organization={IEEE}
}

@inproceedings{liu2024fatformer,
  title={Forgery-aware adaptive transformer for generalizable synthetic image detection},
  author={Liu, Huan and Tan, Zichang and Tan, Chuangchuang and Wei, Yunchao and Wang, Jingdong and Zhao, Yao},
  booktitle={2024 IEEE/CVF Conference on Computer Vision and Pattern Recognition (CVPR)},
  pages={10770--10780},
  year={2024},
  organization={IEEE}
}

@inproceedings{tan2024frequencyFreqNet,
  title={Frequency-aware deepfake detection: Improving generalizability through frequency space domain learning},
  author={Tan, Chuangchuang and Zhao, Yao and Wei, Shikui and Gu, Guanghua and Liu, Ping and Wei, Yunchao},
  booktitle={Proceedings of the AAAI Conference on Artificial Intelligence},
  volume={38},
  number={5},
  pages={5052--5060},
  year={2024}
}

@inproceedings{tan2024rethinkingNPR,
  title={Rethinking the up-sampling operations in cnn-based generative network for generalizable deepfake detection},
  author={Tan, Chuangchuang and Zhao, Yao and Wei, Shikui and Gu, Guanghua and Liu, Ping and Wei, Yunchao},
  booktitle={Proceedings of the IEEE/CVF conference on computer vision and pattern recognition},
  pages={28130--28139},
  year={2024}
}

@inproceedings{koutlis2024RINE,
  title={Leveraging representations from intermediate encoder-blocks for synthetic image detection},
  author={Koutlis, Christos and Papadopoulos, Symeon},
  booktitle={Proceedings of the European Conference on Computer Vision},
  pages={394--411},
  year={2024}
}

@inproceedings{yan2025sanityAIDE,
  title={A sanity check for ai-generated image detection},
  author={Yan, Shilin and Li, Ouxiang and Cai, Jiayin and Hao, Yanbin and Jiang, Xiaolong and Hu, Yao and Xie, Weidi},
  booktitle={International Conference on Learning Representations},
  volume={2025},
  pages={70702--70720},
  year={2025}
}

@inproceedings{tan2025c2p,
  title={C2p-clip: Injecting category common prompt in clip to enhance generalization in deepfake detection},
  author={Tan, Chuangchuang and Tao, Renshuai and Liu, Huan and Gu, Guanghua and Wu, Baoyuan and Zhao, Yao and Wei, Yunchao},
  booktitle={Proceedings of the AAAI Conference on Artificial Intelligence},
  volume={39},
  number={7},
  pages={7184--7192},
  year={2025}
}

@inproceedings{li2025improvingSAFE,
  title={Improving synthetic image detection towards generalization: An image transformation perspective},
  author={Li, Ouxiang and Cai, Jiayin and Hao, Yanbin and Jiang, Xiaolong and Hu, Yao and Feng, Fuli},
  booktitle={Proceedings of the 31st ACM SIGKDD conference on knowledge discovery and data mining v. 1},
  pages={2405--2414},
  year={2025}
}

@inproceedings{yan2025effort,
title={Orthogonal Subspace Decomposition for Generalizable {AI}-Generated Image Detection},
author={Yan, Zhiyuan and Wang, Jiangming and Wang, Zhendong and Jin, Peng and Zhang, Ke-Yue and Chen, Shen and Yao, Taiping and Ding, Shouhong and Wu, Baoyuan and Yuan, Li},
booktitle={Proceedings of the International Conference on Machine Learning},
year={2025}
}

@article{guo2025omniaid,
  title={Omniaid: Decoupling semantic and artifacts for universal ai-generated image detection in the wild},
  author={Guo, Yuncheng and Ye, Junyan and Zhang, Chenjue and Kang, Hengrui and Fu, Haohuan and He, Conghui and Li, Weijia},
  journal={Proceedings of the International Conference on Machine Learning},
  year={2026}
}

@inproceedings{PGCzhou26h,
  title = 	 {{PGC}: Peak-Guided Calibration for Generalizable {AI}-Generated Image Detection},
  author =       {Zhou, Xiaoyu and Fei, Jianwei and Yu, Peipeng and Xie, Jingchang and Cheng, Chong and Xia, Zhihua},
  booktitle = 	 {Proceedings of International Conference on Machine Learning},
  year = 	 {2026},
  }

@inproceedings{li2026towardsIAPL,
  title={Towards generalizable ai-generated image detection via image-adaptive prompt learning},
  author={Li, Yiheng and Tan, Zichang and Xu, Guoqing and Lei, Zhen and Zhou, Xu and Yang, Yang},
  booktitle={Proceedings of the IEEE/CVF Conference on Computer Vision and Pattern Recognition},
  pages={21262--21272},
  year={2026}
}

@article{zhou2026simplicity,
  title={Simplicity prevails: The emergence of generalizable aigi detection in visual foundation models},
  author={Zhou, Yue and He, Xinan and Lin, Kaiqing and Fan, Bing and Ding, Feng and Li, Bin},
  journal={arXiv preprint arXiv:2602.01738},
  year={2026}
}

@article{li2026spherevideo,
  title={{SphereVideo}: Prototype-anchored Hyperspherical Boundary for Continual {AI}-generated Video Detection},
  author={Li, Fei and Yu, Yue and Wang, Yuran and Li, Xinghan and Chen, Jingjing and Jiang, Yu-Gang},
  journal={arXiv preprint arXiv:2608.01334},
  year={2026}
}

@inproceedings{wang2020cnnspot,
  title={{CNN}-generated images are surprisingly easy to spot... for now},
  author={Wang, Sheng-Yu and Wang, Oliver and Zhang, Richard and Owens, Andrew and Efros, Alexei A},
  booktitle={Proceedings of the IEEE/CVF Conference on Computer Vision and Pattern Recognition},
  pages={8695--8704},
  year={2020}
}

@inproceedings{tan2023LGrad,
  title={{Learning on Gradients}: Generalized artifacts representation for gan-generated images detection},
  author={Tan, Chuangchuang and Zhao, Yao and Wei, Shikui and Gu, Guanghua and Wei, Yunchao},
  booktitle={Proceedings of the IEEE/CVF Conference on Computer Vision and Pattern Recognition},
  pages={12105--12114},
  year={2023}
}

@inproceedings{chen2024drct,
  title={DRCT: Diffusion Reconstruction Contrastive Training towards Universal Detection of Diffusion Generated Images.},
  author={Chen, Baoying and Zeng, Jishen and Yang, Jianquan and Yang, Rui},
  booktitle={Proceedings of the International Conference on Machine Learning},
  volume={2},
  number={4},
  pages={6},
  year={2024}
}

@article{zhou2026breakingOMAT,
  title={Breaking latent prior bias in detectors for generalizable aigc image detection},
  author={Zhou, Yue and He, Xinan and Lin, KaiQing and Fan, Bing and Ding, Feng and Li, Bin},
  journal={Advances in Neural Information Processing Systems},
  volume={38},
  pages={30649--30679},
  year={2025}
}

@inproceedings{zhang2025towardsVIB,
  title={Towards universal ai-generated image detection by variational information bottleneck network},
  author={Zhang, Haifeng and He, Qinghui and Bi, Xiuli and Li, Weisheng and Liu, Bo and Xiao, Bin},
  booktitle={2025 IEEE/CVF Conference on Computer Vision and Pattern Recognition (CVPR)},
  pages={23828--23837},
  year={2025},
  organization={IEEE}
}

@inproceedings{deng2009imagenet,
  title={{ImageNet}: A Large-Scale Hierarchical Image Database},
  author={Deng, Jia and Dong, Wei and Socher, Richard and Li, Li-Jia and Li, Kai and Fei-Fei, Li},
  booktitle={2009 IEEE Conference on Computer Vision and Pattern Recognition},
  pages={248--255},
  year={2009}
}

@article{zhu2023genimage,
  title={Genimage: A million-scale benchmark for detecting ai-generated image},
  author={Zhu, Mingjian and Chen, Hanting and Yan, Qiangyu and Huang, Xudong and Lin, Guanyu and Li, Wei and Tu, Zhijun and Hu, Hailin and Hu, Jie and Wang, Yunhe},
  journal={Advances in neural information processing systems},
  volume={36},
  pages={77771--77782},
  year={2023}
}

@article{zhong2023patchcraftAIGD,
  title={Patchcraft: Exploring texture patch for efficient ai-generated image detection},
  author={Zhong, Nan and Xu, Yiran and Li, Sheng and Qian, Zhenxing and Zhang, Xinpeng},
  journal={arXiv preprint arXiv:2311.12397},
  year={2023}
}

@inproceedings{zhou2025holmes,
  title={{AIGI-Holmes}: Towards explainable and generalizable ai-generated image detection via multimodal large language models},
  author={Zhou, Ziyin and Luo, Yunpeng and Wu, Yuanchen and Sun, Ke and Ji, Jiayi and Yan, Ke and Ding, Shouhong and Sun, Xiaoshuai and Wu, Yunsheng and Ji, Rongrong},
  booktitle={Proceedings of the IEEE/CVF International Conference on Computer Vision},
  pages={18746--18758},
  year={2025}
}

@article{chen2025taskAIGInow,
  title={Task-Model Alignment: A Simple Path to Generalizable AI-Generated Image Detection},
  author={Chen, Ruoxin and Gao, Jiahui and Lin, Kaiqing and Zhang, Keyue and Zhao, Yandan and Guan, Isabel and Yao, Taiping and Ding, Shouhong},
  journal={arXiv e-prints},
  pages={arXiv--2512},
  year={2025}
}

@article{wildRF,
  title={Real-time deepfake detection in the real-world},
  author={Cavia, Bar and Horwitz, Eliahu and Reiss, Tal and Hoshen, Yedid},
  journal={arXiv preprint arXiv:2406.09398},
  year={2024}
}

@inproceedings{CommunityAI,
  title={Is artificial intelligence generated image detection a solved problem?},
  author={Li, Ziqiang and Yan, Jiazhen and He, Ziwen and Zeng, Kai and Jiang, Weiwei and Xiong, Lizhi and Fu, Zhangjie},
  booktitle={Advances in Neural information processing systems},
  volume={38},
  year={2026}
}

@article{brock2018largeBiggan,
  title={Large scale GAN training for high fidelity natural image synthesis},
  author={Brock, Andrew and Donahue, Jeff and Simonyan, Karen},
  journal={arXiv preprint arXiv:1809.11096},
  year={2018}
}

@article{nichol2021glide,
  title={Glide: Towards photorealistic image generation and editing with text-guided diffusion models},
  author={Nichol, Alex and Dhariwal, Prafulla and Ramesh, Aditya and Shyam, Pranav and Mishkin, Pamela and McGrew, Bob and Sutskever, Ilya and Chen, Mark},
  journal={arXiv preprint arXiv:2112.10741},
  year={2021}
}

@inproceedings{gu2022vectorVQDM,
  title={Vector quantized diffusion model for text-to-image synthesis},
  author={Gu, Shuyang and Chen, Dong and Bao, Jianmin and Wen, Fang and Zhang, Bo and Chen, Dongdong and Yuan, Lu and Guo, Baining},
  booktitle={2022 IEEE/CVF conference on computer vision and pattern recognition (CVPR)},
  pages={10686--10696},
  year={2022},
  organization={IEEE}
}

@inproceedings{dhariwal2021diffusionADM,
  title={Diffusion models beat gans on image synthesis},
  author={Dhariwal, Prafulla and Nichol, Alexander},
  booktitle={Advances in Neural Information Processing Systems},
  volume={34},
  pages={8780--8794},
  year={2021}
}

@misc{midjourney2022Midjourney,
  author       = {{Midjourney, Inc.}},
  title        = {Midjourney},
  year         = {2022},
  howpublished = {\url{https://www.midjourney.com/home/}},
  note         = {Accessed: Nov. 11, 2026}
}

@misc{unsplash,
  title        = {{Unsplash}},
  howpublished = {\url{https://unsplash.com}}
}

@misc{artstation,
  title        = {{ArtStation}},
  howpublished = {\url{https://www.artstation.com}}
}

@misc{civitai,
  title        = {{Civitai}},
  howpublished = {\url{https://civitai.com}}
}

@misc{liblib,
  title        = {{Liblib}},
  howpublished = {\url{https://www.liblib.art}}
}

@misc{wukong2022,
  author       = {{Huawei MindSpore}},
  title        = {Wukong},
  year         = {2022},
  howpublished = {\url{https://xihe.mindspore.cn/modelzoo/wukong}},
  note         = {Accessed: Nov. 11, 2026}
}

@misc{blackforestlabs2024flux,
  author = {{Black Forest Labs}},
  title  = {{FLUX}},
  year   = {2024},
  howpublished = {\url{https://github.com/black-forest-labs/flux}}
}

@article{chen2025januspro,
  author  = {Chen, Xiaokang and Wu, Zhiyu and Liu, Xingchao and Pan, Zizheng
             and Liu, Wen and Xie, Zhenda and Yu, Xingkai and Ruan, Chong},
  title   = {{Janus-Pro}: Unified Multimodal Understanding and Generation
             with Data and Model Scaling},
  journal = {arXiv preprint arXiv:2501.17811},
  year    = {2025}
}

@article{sun2024llamagen,
  author  = {Sun, Peize and Jiang, Yi and Chen, Shoufa and Zhang, Shilong
             and Peng, Bingyue and Luo, Ping and Yuan, Zehuan},
  title   = {Autoregressive Model Beats Diffusion: {Llama} for Scalable
             Image Generation},
  journal = {arXiv preprint arXiv:2406.06525},
  year    = {2024}
}

@inproceedings{chen2024pixartsigma,
  title={Pixart-$\sigma$: Weak-to-strong training of diffusion transformer for 4k text-to-image generation},
  author={Chen, Junsong and Ge, Chongjian and Xie, Enze and Wu, Yue and Yao, Lewei and Ren, Xiaozhe and Wang, Zhongdao and Luo, Ping and Lu, Huchuan and Li, Zhenguo},
  booktitle={European Conference on Computer Vision},
  pages={74--91},
  year={2024},
}

@inproceedings{esser2024scaling,
  author    = {Esser, Patrick and Kulal, Sumith and Blattmann, Andreas
               and Entezari, Rahim and M{\"u}ller, Jonas and Saini, Harry
               and Levi, Yam and Lorenz, Dominik and Sauer, Axel
               and Boesel, Frederic and Podell, Dustin and Dockhorn, Tim
               and English, Zion and Rombach, Robin},
  title     = {Scaling Rectified Flow Transformers for High-Resolution
               Image Synthesis},
  booktitle = {Proceedings of the 41st International Conference on Machine Learning},
  pages     = {12606--12633},
  volume    = {235},
  year      = {2024}
}

@inproceedings{xie2024showo,
title={Show-o: One single transformer to unify multimodal understanding and generation},
  author={Xie, Jinheng and Mao, Weijia and Bai, Zechen and Zhang, David Junhao and Wang, Weihao and Lin, Kevin Qinghong and Gu, Yuchao and Chen, Zhijie and Yang, Zhenheng and Shou, Mike Zheng},
  booktitle={International Conference on Learning Representations},
  volume={2025},
  pages={28240--28264},
  year={2025}
}

@inproceedings{tian2024var,
  title={Visual autoregressive modeling: Scalable image generation via next-scale prediction},
  author={Tian, Keyu and Jiang, Yi and Yuan, Zehuan and Peng, Bingyue and Wang, Liwei},
  booktitle={Advances in neural information processing systems},
  volume={37},
  pages={84839--84865},
  year={2024}
}

@misc{fortin2025geminiflashimage,
  author       = {Fortin, Alisa and Vernade, Guillaume and Kampf, Kat and Reshi, Ammaar},
  title        = {Introducing {Gemini 2.5 Flash Image}, Our State-of-the-Art Image Model},
  howpublished = {Google Developers Blog, \url{https://developers.googleblog.com/en/introducing-gemini-2-5-flash-image/}},
  year         = {2025}
}

@article{openai2025gpt4oimage,
  title={Gpt-4o system card},
  author={Hurst, Aaron and Lerer, Adam and Goucher, Adam P and Perelman, Adam and Ramesh, Aditya and Clark, Aidan and Ostrow, AJ and Welihinda, Akila and Hayes, Alan and Radford, Alec and others},
  journal={arXiv preprint arXiv:2410.21276},
  year={2024}
}

@misc{volcengine2025jimeng,
  author = {{Volcengine}},
  title  = {{Jimeng AI} Opens {API} Access Through {Volcengine}},
  year   = {2025},
  howpublished = {\url{https://developer.volcengine.com/articles/7546180861461266483}}
}

@article{kuaishou2025kling2,
  title={Klingavatar 2.0 technical report},
  author={Team, Kling and Chen, Jialu and Ding, Yikang and Fang, Zhixue and Gai, Kun and Gao, Yuan and He, Kang and Hua, Jingyun and Jiang, Boyuan and Lao, Mingming and others},
  journal={arXiv preprint arXiv:2512.13313},
  year={2025}
}

@article{minimax2025image01,
  title={Minimax-01: Scaling foundation models with lightning attention},
  author={Li, Aonian and Gong, Bangwei and Yang, Bo and Shan, Boji and Liu, Chang and Zhu, Cheng and Zhang, Chunhao and Guo, Congchao and Chen, Da and Li, Dong and others},
  journal={arXiv preprint arXiv:2501.08313},
  year={2025}
}

@misc{bfl2024flux11pro,
  author = {{Black Forest Labs}},
  title  = {Announcing {FLUX1.1 [pro]} and the {BFL API}},
  year   = {2024},
  howpublished = {\url{https://bfl.ai/blog/24-10-02-flux}}
}

@misc{bfl2025fluxkreadev,
  author = {{Black Forest Labs}},
  title  = {{FLUX.1 Krea [dev]} Model Card},
  year   = {2025},
  howpublished = {\url{https://huggingface.co/black-forest-labs/FLUX.1-Krea-dev}}
}

@misc{bfl2024fluxdev,
  author = {{Black Forest Labs}},
  title  = {{FLUX.1 [dev]} Model Card},
  year   = {2024},
  howpublished = {\url{https://huggingface.co/black-forest-labs/FLUX.1-dev}}
}

@misc{bfl2025fluxkontextdev,
  author = {{Black Forest Labs}},
  title  = {{FLUX.1 Kontext [dev]} Model Card},
  year   = {2025},
  howpublished = {\url{https://huggingface.co/black-forest-labs/FLUX.1-Kontext-dev}}
}

@inproceedings{zhu2017cyclegan,
  author    = {Zhu, Jun-Yan and Park, Taesung and Isola, Phillip and Efros, Alexei A.},
  title     = {Unpaired Image-to-Image Translation Using Cycle-Consistent Adversarial Networks},
  booktitle = {Proceedings of the IEEE International Conference on Computer Vision},
  pages     = {2223--2232},
  year      = {2017}
}

@inproceedings{choi2018stargan,
  author    = {Choi, Yunjey and Choi, Minje and Kim, Munyoung and Ha, Jung-Woo
               and Kim, Sunghun and Choo, Jaegul},
  title     = {{StarGAN}: Unified Generative Adversarial Networks for Multi-Domain
               Image-to-Image Translation},
  booktitle = {Proceedings of the IEEE Conference on Computer Vision and Pattern Recognition},
  pages     = {8789--8797},
  year      = {2018}
}

@inproceedings{park2019gaugan,
  author    = {Park, Taesung and Liu, Ming-Yu and Wang, Ting-Chun and Zhu, Jun-Yan},
  title     = {Semantic Image Synthesis with Spatially-Adaptive Normalization},
  booktitle = {Proceedings of the IEEE/CVF Conference on Computer Vision and Pattern Recognition},
  pages     = {2337--2346},
  year      = {2019}
}

@inproceedings{rossler2019faceforensics,
  author    = {R{\"o}ssler, Andreas and Cozzolino, Davide and Verdoliva, Luisa
               and Riess, Christian and Thies, Justus and Nie{\ss}ner, Matthias},
  title     = {{FaceForensics++}: Learning to Detect Manipulated Facial Images},
  booktitle = {Proceedings of the IEEE/CVF International Conference on Computer Vision},
  pages     = {1--11},
  year      = {2019}
}

@inproceedings{chen2018sitd,
  author    = {Chen, Chen and Chen, Qifeng and Xu, Jia and Koltun, Vladlen},
  title     = {Learning to See in the Dark},
  booktitle = {Proceedings of the IEEE Conference on Computer Vision and Pattern Recognition},
  pages     = {3291--3300},
  year      = {2018}
}

@inproceedings{dai2019san,
  author    = {Dai, Tao and Cai, Jianrui and Zhang, Yongbing and Xia, Shu-Tao and Zhang, Lei},
  title     = {Second-Order Attention Network for Single Image Super-Resolution},
  booktitle = {Proceedings of the IEEE/CVF Conference on Computer Vision and Pattern Recognition},
  pages     = {11065--11074},
  year      = {2019}
}

@inproceedings{chen2017crn,
  author    = {Chen, Qifeng and Koltun, Vladlen},
  title     = {Photographic Image Synthesis with Cascaded Refinement Networks},
  booktitle = {Proceedings of the IEEE International Conference on Computer Vision},
  pages     = {1511--1520},
  year      = {2017}
}

@inproceedings{li2019conditionalimle,
  author    = {Li, Ke and Zhang, Tianhao and Malik, Jitendra},
  title     = {Diverse Image Synthesis from Semantic Layouts via Conditional {IMLE}},
  booktitle = {Proceedings of the IEEE/CVF International Conference on Computer Vision},
  pages     = {4220--4229},
  year      = {2019}
}

@misc{dayma2021dallemini,
  author       = {Dayma, Boris and Patil, Suraj and Cuenca, Pedro and Saifullah, Khalid
                  and Abraham, Tanishq and L{\^e} Kh{\'{\u{a}}}c, Ph{\'u}c
                  and Melas, Luke and Ghosh, Ritobrata},
  title        = {{DALL-E Mini}},
  year         = {2021},
  doi          = {10.5281/zenodo.5146400},
  howpublished = {\url{https://github.com/borisdayma/dalle-mini}}
}

@inproceedings{karras2020stylegan2,
  author    = {Karras, Tero and Laine, Samuli and Aittala, Miika and Hellsten, Janne
               and Lehtinen, Jaakko and Aila, Timo},
  title     = {Analyzing and Improving the Image Quality of {StyleGAN}},
  booktitle = {Proceedings of the IEEE/CVF Conference on Computer Vision and Pattern Recognition},
  pages     = {8110--8119},
  year      = {2020}
}

@misc{west2023whichfaceisreal,
  author       = {West, Jevin and Bergstrom, Carl},
  title        = {Which Face Is Real?},
  howpublished = {\url{https://www.whichfaceisreal.com/}},
  year         = {2023}
}

@article{podell2023sdxl,
  author  = {Podell, Dustin and English, Zion and Lacey, Kyle and Blattmann, Andreas
             and Dockhorn, Tim and M{\"u}ller, Jonas and Penna, Joe and Rombach, Robin},
  title   = {{SDXL}: Improving Latent Diffusion Models for High-Resolution Image Synthesis},
  journal = {arXiv preprint arXiv:2307.01952},
  year    = {2023}
}
}

\clearpage
\appendix
\suppressfloats[t]
\section*{Appendix}
\section{Extended introduction}

\begin{figure}[!ht]
  \centering
  \includegraphics[width=\columnwidth]{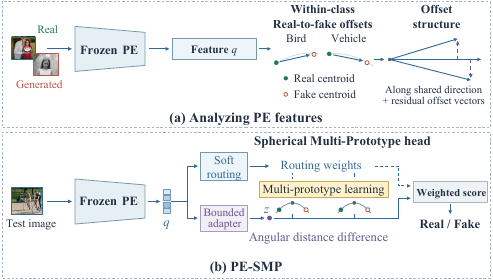}
  \caption{\textbf{Motivation and design of the SMP head.}
    \textbf{(a)} We use the image feature $q$ to estimate class-wise
    real-to-fake offsets and decompose each offset into its component
    along the shared direction and a residual offset vector.
    \textbf{(b)} PE-SMP uses the same feature $q$ for soft routing and
    bounded adaptation. The head scores the adapted feature $z$ against
    the context-specific prototypes from multi-prototype learning.}
  \label{fig:smp-appendix-overview}
\end{figure}

The introduction examines how features extracted by a frozen Perception
Encoder (PE)~\cite{bolya2026perceptionPEcore} shift from real to generated
images within the same semantic class. \Cref{fig:smp-appendix-overview}(a)
traces this analysis from within-class real-to-fake offsets to their
alignment across classes.
The offsets exhibit a shared direction. After removing the component of
each offset along that direction, the residual offset vectors align more
closely within semantic groups. These observations motivate a shared
offset across visual contexts and a local offset for each context.

Our final detector, PE-SMP, pairs a frozen PE backbone with a Spherical
Multi-Prototype (SMP) head, as shown in \cref{fig:smp-appendix-overview}(b).
Through multi-prototype learning, the head learns one real and one fake
prototype per visual context using shared and local offsets. For each
image, soft routing uses $q$ to weight the contexts, and the
bounded adapter maps $q$ to $z$. The detector computes the difference
between the angular distances from $z$ to the real and fake prototypes
in each context. It aggregates these differences using the routing
weights to obtain the detection score.

The SMP design motivated by the PE feature analysis also improves
detection with other pretrained backbones. We train separate SMP
heads on frozen CLIP~\cite{radford2021learningclip} and
\mbox{DINOv3}~\cite{zhou2026simplicity} features and compare each with a
linear head trained on the same features. Across the five tested
configurations of PE, CLIP, and DINOv3, SMP improves mean accuracy over
the corresponding linear head (\cref{tab:ablation_backbone_heads}).

\subsection{Cross-class offset alignment}
\label{sec:smp-offset-alignment}

We analyze the 1,000 ImageNet classes in the GenImage SD v1.4 training
split~\cite{zhu2023genimage}. We use the same unit-normalized image
feature $q$ as PE-SMP, obtained from the block-50 patch tokens of
PE-Core G/14. For each class $c$, we separately average the real and
fake features and normalize the resulting centroids to obtain
$\mu_c^r$ and $\mu_c^f$. The real-to-fake offset vector is
$\delta_c = \operatorname{Log}_{\mu_c^r}(\mu_c^f)$, where
$\operatorname{Log}$ is the spherical logarithm map. This vector is
tangent to the sphere at $\mu_c^r$, and its norm equals the angle between
the two centroids (\cref{fig:smp-offset-definition}, left).
\Cref{fig:smp-class-projection} illustrates these offsets for wheeled
vehicles, mammals, and birds.

\Cref{fig:smp-offset-angle} shows the distribution of offset magnitudes
across classes. \Cref{fig:smp-offset-cosine} shows the distribution of
cosine similarities between the offset vectors of distinct classes.
Both panels compare the real/fake split with size-matched random splits
within each class.

We use the direction of the mean offset $\bar{\delta}$ in the ambient
feature space as the shared direction. As illustrated in
\cref{fig:smp-offset-definition} (right), removing the component of
$\delta_c$ along the shared direction gives the orthogonal residual
offset vector
\begin{equation}
  r_c = \delta_c
    - \frac{\delta_c^\top \bar{\delta}}{\lVert\bar{\delta}\rVert_2^2}
      \bar{\delta}.
\end{equation}

\Cref{fig:smp-offset-residuals} compares the directions of the residual
offset vectors estimated from separate image halves. Each cell reports
the mean cosine similarity across pairs of distinct classes. Diagonal
cells compare classes within the same semantic group, and off-diagonal
cells compare classes from different groups.

\section{Evaluation datasets}

We describe the evaluation datasets for the GenImage, UniversalFakeDetect,
and AIGCDetectBenchmark settings, including their sources and the numbers of
real and generated images in each test set.

\subsection{GenImage setting}

Following the GenImage protocol~\cite{zhu2023genimage,PGCzhou26h,zhou2026simplicity},
we train PE-SMP only on the SD v1.4 training split and use the same
detector on all \textbf{seven benchmarks}. \Cref{tab:genimage_setting_counts}
lists the test subsets and their class counts.

\begin{table*}[!t]
\centering
\caption{\textbf{Test-set composition for the seven detection benchmarks in the GenImage setting.}
  For AIGI-Now, we group the 18 test subsets into nine rows by generator,
  combining the pixel-level and semantic-level counts in each row.
  CommunityAI and SocialRF counts refer to the evaluated splits.}
\label{tab:genimage_setting_counts}
\small
\setlength{\aboverulesep}{0pt}
\setlength{\belowrulesep}{0pt}
\renewcommand{\arraystretch}{1.08}
\begin{tabular*}{\textwidth}{@{}l|lll@{\extracolsep{\fill}\hspace{2\tabcolsep}}rr@{}}
  \toprule
  Scenario & Benchmark & No. of subsets & Test subset & Real images & Fake images \\
  \midrule
  \multirow{9}{*}{Standard}
    & \multirow{9}{*}{GenImage~\cite{zhu2023genimage}}
    & \multirow{9}{*}{8} & Midjourney~\cite{midjourney2022Midjourney} & 6,000 & 6,000 \\
    & & & SD v1.4~\cite{rombach2022highLDM} & 6,000 & 6,000 \\
    & & & SD v1.5~\cite{rombach2022highLDM} & 8,000 & 8,000 \\
    & & & ADM~\cite{dhariwal2021diffusionADM} & 6,000 & 6,000 \\
    & & & GLIDE~\cite{nichol2021glide} & 6,000 & 6,000 \\
    & & & Wukong~\cite{wukong2022} & 6,000 & 6,000 \\
    & & & VQDM~\cite{gu2022vectorVQDM} & 6,000 & 6,000 \\
    & & & BigGAN~\cite{brock2018largeBiggan} & 6,000 & 6,000 \\
  \cmidrule(l){4-6}
    & & & Total & 50,000 & 50,000 \\
  \midrule
  \multirow{7}{*}{In-the-wild}
    & Chameleon~\cite{yan2025sanityAIDE} & 1 & -- & 14,863 & 11,170 \\
    & CommunityAI~\cite{CommunityAI} & 1 & -- & 6,000 & 6,000 \\
    & SocialRF~\cite{CommunityAI} & 1 & -- & 3,000 & 3,000 \\
  \cmidrule{2-6}
    & \multirow{4}{*}{WildRF~\cite{wildRF}}
    & \multirow{4}{*}{3} & Reddit & 750 & 750 \\
    & & & Facebook & 160 & 160 \\
    & & & Twitter & 340 & 340 \\
  \cmidrule(l){4-6}
    & & & Total & 1,250 & 1,250 \\
  \midrule
  \multirow{21}{*}{Unseen-generator}
    & \multirow{11}{*}{AIGI-Holmes~\cite{zhou2025holmes}}
    & \multirow{11}{*}{10} & FLUX~\cite{blackforestlabs2024flux} & 5,000 & 5,000 \\
    & & & Infinity~\cite{han2025infinity} & 5,000 & 5,000 \\
    & & & Janus~\cite{wu2025janus} & 5,000 & 5,000 \\
    & & & Janus-Pro-1B~\cite{chen2025januspro} & 5,000 & 5,000 \\
    & & & Janus-Pro-7B~\cite{chen2025januspro} & 5,000 & 5,000 \\
    & & & LlamaGen~\cite{sun2024llamagen} & 5,000 & 5,000 \\
    & & & PixArt-XL~\cite{chen2024pixartsigma} & 5,000 & 5,000 \\
    & & & SD3.5-Large~\cite{esser2024scaling} & 5,000 & 5,000 \\
    & & & Show-o~\cite{xie2024showo} & 5,000 & 5,000 \\
    & & & VAR~\cite{tian2024var} & 5,000 & 5,000 \\
  \cmidrule(l){4-6}
    & & & Total & 50,000 & 50,000 \\
  \cmidrule{2-6}
    & \multirow{10}{*}{AIGI-Now~\cite{chen2025taskAIGInow}}
    & \multirow{10}{*}{9} & Nano Banana~\cite{fortin2025geminiflashimage} & 2,000 & 2,000 \\
    & & & GPT-4o~\cite{openai2025gpt4oimage} & 2,000 & 2,000 \\
    & & & Jimeng~\cite{volcengine2025jimeng} & 2,000 & 2,000 \\
    & & & Kling~\cite{kuaishou2025kling2} & 2,000 & 2,000 \\
    & & & Minimax~\cite{minimax2025image01} & 2,000 & 2,000 \\
    & & & FLUX Pro~\cite{bfl2024flux11pro} & 2,000 & 2,000 \\
    & & & FLUX Krea~\cite{bfl2025fluxkreadev} & 2,000 & 2,000 \\
    & & & FLUX Dev~\cite{bfl2024fluxdev} & 2,000 & 2,000 \\
    & & & FLUX Kontext~\cite{bfl2025fluxkontextdev} & 2,000 & 2,000 \\
  \cmidrule(l){4-6}
    & & & Total & 18,000 & 18,000 \\
  \bottomrule
\end{tabular*}
\end{table*}

\paragraph{Standard scenario.}
\textbf{GenImage}~\cite{zhu2023genimage} draws its real images from
ImageNet~\cite{deng2009imagenet} and uses the same 1,000 class labels to
condition eight image generators spanning diffusion models and GANs.
Each test subset combines images from one generator with a separate set
of real images. The eight subsets contain 50,000 real and 50,000 generated
images in total. Within GenImage, the SD v1.4 subset serves as the in-distribution test set, 
and the other seven subsets allow us to compare performance across generators.

\paragraph{In-the-wild scenario.}
\textbf{Chameleon}~\cite{yan2025sanityAIDE} combines real photographs
from online photography platforms such as Unsplash~\cite{unsplash}
with generated images posted on ArtStation~\cite{artstation},
Civitai~\cite{civitai}, and Liblib~\cite{liblib}.
The benchmark authors used the same search terms for both classes to
collect images with similar subject matter. They retained generated images
that two independent annotators judged photorealistic, covering people,
animals, objects, and scenes.

\textbf{CommunityAI}~\cite{CommunityAI} is an AIGIBench subset whose
generated images come from posts on AI art communities such as
ArtStation, Civitai, and Liblib. The authors remove near-duplicates and
low-quality images.

\textbf{SocialRF}~\cite{CommunityAI} is an AIGIBench subset collected
from social platforms such as Reddit, Facebook, and Twitter. The authors
used photography-related and AI-art hashtags to retrieve real and generated
images, respectively.

\textbf{WildRF}~\cite{wildRF} contains images manually curated from
Reddit, Facebook, and Twitter. Its test split is organized by platform,
with equal numbers of real and fake images in each subset.

\paragraph{Unseen-generator scenario.}
We evaluate generalization to recent image generators using AIGI-Holmes
and AIGI-Now.
For \textbf{AIGI-Holmes}~\cite{zhou2025holmes}, we use ten test subsets,
one per generator, spanning diffusion and autoregressive models. Each
subset contains 5,000 real images from COCO and 5,000 images produced by
the corresponding generator.

\textbf{AIGI-Now} comprises nine generator subsets~\cite{chen2025taskAIGInow}.
Each subset contains a pixel-level split based on original COCO captions
and a semantic-level split based on surreal rewrites of those captions.
The benchmark applies post-processing to both real and fake images in
the semantic-level split to disrupt low-level cues while preserving
semantic inconsistencies. The two splits are equal in size, with 1,000
real and 1,000 fake images each.

\subsection{UniversalFakeDetect setting}

\textbf{UniversalFakeDetect}~\cite{ojha2023towards} is a benchmark with
19 test subsets (\cref{tab:ufd_setting_counts}). Eleven come from
CNNSpot~\cite{wang2020cnnspot} and cover GANs, face swaps, low-light
enhancement (SITD), super-resolution (SAN), and semantic image
synthesis (CRN and IMLE). The remaining eight comprise guided diffusion,
three LDM configurations, three GLIDE configurations, and DALL-E Mini
(labeled DALL-E in the table).
The LDM configurations vary the number of steps and the use of
classifier-free guidance. The GLIDE configurations vary the step
counts in the base and upsampling stages. The real images for LDM,
GLIDE, and DALL-E Mini come from LAION.

We train on the four-class ProGAN~\cite{karras2017progressiveprogan}
split covering cars, cats, chairs, and horses, with real images from LSUN.

\subsection{AIGCDetectBenchmark setting}

\textbf{AIGCDetectBenchmark}~\cite{zhong2023patchcraftAIGD} is a benchmark
with 17 test subsets (\cref{tab:aigcdetect_setting_counts}). Seven GAN
subsets and Which Face Is Real (WFIR) use fake images from
CNNSpot~\cite{wang2020cnnspot}. WFIR contains StyleGAN-generated faces and
real portraits from FFHQ~\cite{west2023whichfaceisreal}. Seven other
subsets come from GenImage~\cite{zhu2023genimage}, and the benchmark
authors curated the DALL-E 2 and SDXL subsets.

Following the evaluation setup in OmniAID~\cite{guo2025omniaid}, we train
on the ProGAN~\cite{karras2017progressiveprogan} split covering 20 semantic
categories, with real images from LSUN.

\newpage
\noindent
\begin{minipage}{\columnwidth}
\centering
\captionsetup{type=table}
\caption{\textbf{Test-set composition of the UniversalFakeDetect benchmark.}
  Each row gives the real and fake counts for an evaluated subset.}
\label{tab:ufd_setting_counts}
\small
\begin{tabular*}{\linewidth}{@{\extracolsep{\fill}}lrr@{}}
  \toprule
  Test subset & Real images & Fake images \\
  \midrule
  ProGAN~\cite{karras2017progressiveprogan} & 4,000 & 4,000 \\
  CycleGAN~\cite{zhu2017cyclegan} & 1,321 & 1,321 \\
  BigGAN~\cite{brock2018largeBiggan} & 2,000 & 2,000 \\
  StyleGAN~\cite{karras2019stylegan} & 5,991 & 5,991 \\
  GauGAN~\cite{park2019gaugan} & 5,000 & 5,000 \\
  StarGAN~\cite{choi2018stargan} & 1,999 & 1,999 \\
  DeepFakes~\cite{rossler2019faceforensics} & 2,707 & 2,698 \\
  SITD~\cite{chen2018sitd} & 180 & 180 \\
  SAN~\cite{dai2019san} & 219 & 219 \\
  CRN~\cite{chen2017crn} & 6,382 & 6,382 \\
  IMLE~\cite{li2019conditionalimle} & 6,382 & 6,382 \\
  Guided diffusion~\cite{dhariwal2021diffusionADM} & 1,000 & 1,000 \\
  LDM 200 steps~\cite{rombach2022highLDM} & 1,000 & 1,000 \\
  LDM 200 steps with CFG~\cite{rombach2022highLDM} & 1,000 & 1,000 \\
  LDM 100 steps~\cite{rombach2022highLDM} & 1,000 & 1,000 \\
  GLIDE 100/27~\cite{nichol2021glide} & 1,000 & 1,000 \\
  GLIDE 50/27~\cite{nichol2021glide} & 1,000 & 1,000 \\
  GLIDE 100/10~\cite{nichol2021glide} & 1,000 & 1,000 \\
  DALL-E~\cite{dayma2021dallemini} & 1,000 & 1,000 \\
  \midrule
  Total & 44,181 & 44,172 \\
  \bottomrule
\end{tabular*}
\end{minipage}
\par\medskip
\noindent
\begin{minipage}{\columnwidth}
\centering
\captionsetup{type=table}
\caption{\textbf{Test-set composition of AIGCDetectBenchmark.}
  The counts are from the evaluated release used for \cref{tab:aigcdetect}.}
\label{tab:aigcdetect_setting_counts}
\small
\begin{tabular*}{\linewidth}{@{\extracolsep{\fill}}lrr@{}}
  \toprule
  Test subset & Real images & Fake images \\
  \midrule
  ProGAN~\cite{karras2017progressiveprogan} & 4,000 & 4,000 \\
  StyleGAN~\cite{karras2019stylegan} & 5,991 & 5,991 \\
  BigGAN~\cite{brock2018largeBiggan} & 2,000 & 2,000 \\
  CycleGAN~\cite{zhu2017cyclegan} & 1,321 & 1,321 \\
  StarGAN~\cite{choi2018stargan} & 1,999 & 1,999 \\
  GauGAN~\cite{park2019gaugan} & 5,000 & 5,000 \\
  StyleGAN2~\cite{karras2020stylegan2} & 7,988 & 7,988 \\
  WFIR~\cite{west2023whichfaceisreal} & 1,000 & 1,000 \\
  ADM~\cite{dhariwal2021diffusionADM} & 6,000 & 6,000 \\
  GLIDE~\cite{nichol2021glide} & 6,000 & 6,000 \\
  Midjourney~\cite{midjourney2022Midjourney} & 6,000 & 6,000 \\
  SD v1.4~\cite{rombach2022highLDM} & 6,000 & 6,000 \\
  SD v1.5~\cite{rombach2022highLDM} & 8,000 & 8,000 \\
  VQDM~\cite{gu2022vectorVQDM} & 6,000 & 6,000 \\
  Wukong~\cite{wukong2022} & 6,000 & 6,000 \\
  DALL-E 2~\cite{ramesh2022DALL-E3} & 1,000 & 1,000 \\
  SDXL~\cite{podell2023sdxl} & 2,000 & 2,000 \\
  \midrule
  Total & 76,299 & 76,299 \\
  \bottomrule
\end{tabular*}
\end{minipage}
\par
\begin{table*}[!t]
  \centering
  \caption{\textbf{Cross-benchmark evaluation on seven detection benchmarks using Average Precision (\%).}
    $^\clubsuit$: retrained using the official code under the same training setting.
    $^\spadesuit$: evaluated using the publicly released weights.
    \textbf{Following the GenImage protocol, all models are trained solely on the SDv1.4 subset and then evaluated on each benchmark to assess cross-dataset generalization.}
    The best and second-best results are in \best{bold} and \second{underlined}, respectively.}
  \label{tab:cross_bench_ap}
  \setlength{\tabcolsep}{5.85pt}
  \resizebox{\textwidth}{!}{%
  \begin{tabular}{lc|ccccccc|c}
    \toprule
    Methods & Venues & GenImage & Chameleon & CommunityAI & SocialRF & WildRF & AIGI-Holmes & AIGI-Now & Avg. \\
    \midrule
    NPR$^\spadesuit$~\cite{tan2024rethinkingNPR} & CVPR 2024 & 95.82 & 54.43 & 67.09 & 60.96 & 66.37 & 91.69 & 78.66 & 73.57 \\
    DRCT$^\spadesuit$~\cite{chen2024drct} & ICML 2024 & 98.32 & 69.74 & 85.93 & \second{78.35} & \second{86.20} & 96.19 & 76.86 & \second{84.51} \\
    OMAT$^\spadesuit$~\cite{zhou2026breakingOMAT} & NeurIPS 2025 & 98.68 & 68.81 & 72.81 & 66.60 & 78.40 & 95.02 & 67.39 & 78.24 \\
    SAFE$^\clubsuit$~\cite{li2025improvingSAFE} & KDD 2025 & 95.08 & 43.91 & 59.87 & 62.10 & 70.48 & 87.54 & 69.92 & 69.84 \\
    AIDE$^\spadesuit$~\cite{yan2025sanityAIDE} & ICLR 2025 & 97.76 & 66.55 & 78.60 & 67.08 & 72.55 & 97.69 & \second{93.73} & 81.99 \\
    Effort$^\spadesuit$~\cite{yan2025effort} & ICML 2025 & 99.45 & 64.77 & 70.70 & 67.22 & 81.11 & 96.00 & 90.04 & 81.33 \\
    C2P-CLIP$^\spadesuit$~\cite{tan2025c2p} & AAAI 2025 & 99.04 & 51.22 & 57.66 & 56.13 & 64.97 & 98.78 & 93.50 & 74.47 \\
    VIB-Net$^\spadesuit$~\cite{zhang2025towardsVIB} & CVPR 2025 & 96.09 & 55.14 & 59.61 & 61.96 & 72.66 & 97.53 & 88.10 & 75.87 \\
    OmniAID$^\spadesuit$~\cite{guo2025omniaid} & ICML 2026 & 98.57 & \second{85.06} & \second{91.29} & 64.84 & 72.88 & 90.54 & 76.75 & 82.85 \\
    PGC$^\spadesuit$~\cite{PGCzhou26h} & ICML 2026 & \best{99.99} & 61.51 & 74.69 & 75.63 & 85.16 & 98.30 & 85.99 & 83.04 \\
    IAPL$^\spadesuit$~\cite{li2026towardsIAPL} & CVPR 2026 & 99.46 & 64.22 & 72.61 & 61.00 & 64.85 & 62.76 & 59.81 & 69.24 \\
    \midrule
    \rowcolor{oursrow}
    PE-SMP (Ours) & -- & \second{99.87} & \best{99.51} & \best{99.80} & \best{99.15} & \best{99.92} & \best{99.99} & \best{99.74} & \best{99.71} \\
    \bottomrule
  \end{tabular}%
  }
\end{table*}

\newpage

\section{Additional benchmark results}

We complement the main paper's accuracy (Acc) comparisons with average
precision (AP), treating AI-generated images as the positive class. 
In the GenImage setting, we report AP across seven benchmarks and provide
per-subset Acc and AP on GenImage, AIGI-Holmes, and AIGI-Now. In the UniversalFakeDetect setting, we report AP on 19 test
subsets.

\subsection{GenImage setting}

\Cref{tab:cross_bench_ap} reports AP across seven benchmarks for
detectors trained on GenImage SD v1.4. PE-SMP achieves the highest mean
AP (99.71\%) and leads on all in-the-wild and unseen-generator
benchmarks. On GenImage, several methods are close to saturation and PGC
retains a slight lead. The in-the-wild benchmarks separate the detectors
more clearly. On SocialRF, PE-SMP has its largest lead over the strongest
baseline, at 20.8 percentage points. SMP learns a shared real-to-fake
offset and a local adjustment for each prototype pair. Soft routing
emphasizes pairs from contexts similar to the input.
The gains on in-the-wild images are consistent with the benefit of
scoring images against prototype pairs in relevant visual contexts.

\paragraph{GenImage.}
\Cref{tab:genimage,tab:genimage_ap} report Acc and AP on the GenImage
test subsets. PGC leads on both mean metrics, and several detectors
approach saturation on the closely related SD v1.4 and SD v1.5 subsets.
ADM and BigGAN show wider differences among detectors. PE-SMP leads on
BigGAN in both metrics, showing that prototype pairs learned from a
diffusion training source can also separate real images from GAN
outputs. On ADM, PE-SMP retains 99.26\% AP alongside its lowest
fixed-threshold Acc among the GenImage subsets (92.98\%).


\begin{table*}[!t]
\begin{minipage}[t]{\textwidth}
  \centering
  \caption{\textbf{Accuracy (\%) on GenImage~\cite{zhu2023genimage}.}
    Each row uses the same checkpoint as in \cref{tab:cross_bench_ap}.
    $^\clubsuit$ denotes retraining with the official code on SDv1.4, and $^\spadesuit$ denotes evaluation with publicly released weights.
    The best and second-best results are in \best{bold} and \second{underlined}, respectively, based on unrounded scores.}
  \label{tab:genimage}
  \setlength{\tabcolsep}{8pt}
  \resizebox{\textwidth}{!}{%
  \begin{tabular}{lc|cccccccc|c}
    \toprule
    Methods & Venues & Midjourney & SDv1.4 & SDv1.5 & ADM & GLIDE & Wukong & VQDM & BigGAN & Avg. \\
    \midrule
    LGrad$^\clubsuit$~\cite{tan2023LGrad} & CVPR 2023 & 75.33 & 86.17 & 85.88 & 53.59 & 71.53 & 85.06 & 47.27 & 66.61 & 71.43 \\
    NPR$^\spadesuit$~\cite{tan2024rethinkingNPR} & CVPR 2024 & 89.91 & 90.80 & 90.81 & 84.69 & 90.37 & 90.71 & 87.04 & 81.86 & 88.27 \\
    DRCT$^\spadesuit$~\cite{chen2024drct} & ICML 2024 & 94.63 & 99.88 & 99.83 & 61.78 & 65.92 & 99.91 & 74.88 & 59.81 & 82.08 \\
    RINE$^\clubsuit$~\cite{koutlis2024RINE} & ECCV 2024 & 78.78 & 99.93 & 99.81 & 62.49 & 61.66 & 99.86 & 87.93 & 53.38 & 80.48 \\
    OMAT$^\spadesuit$~\cite{zhou2026breakingOMAT} & NeurIPS 2025 & 90.29 & 97.44 & 97.33 & 83.68 & 97.37 & 97.51 & 95.45 & 97.26 & 94.54 \\
    SAFE$^\clubsuit$~\cite{li2025improvingSAFE} & KDD 2025 & 88.40 & 98.32 & 98.21 & 63.49 & 92.56 & 95.66 & 84.21 & 53.63 & 84.31 \\
    AIDE$^\spadesuit$~\cite{yan2025sanityAIDE} & ICLR 2025 & 81.40 & 99.76 & 99.75 & 78.48 & 91.78 & 98.85 & 80.26 & 66.83 & 87.14 \\
    Effort$^\spadesuit$~\cite{yan2025effort} & ICML 2025 & 82.40 & 99.83 & 99.81 & 78.78 & 93.32 & 97.42 & 91.70 & 77.62 & 90.11 \\
    C2P-CLIP$^\spadesuit$~\cite{tan2025c2p} & AAAI 2025 & \second{96.97} & 98.50 & 98.16 & 89.76 & 97.24 & 98.06 & 95.74 & 93.83 & 96.03 \\
    VIB-Net$^\spadesuit$~\cite{zhang2025towardsVIB} & CVPR 2025 & 88.31 & 99.44 & 99.30 & 74.25 & 73.56 & 98.64 & 89.24 & 55.56 & 84.79 \\
    OmniAID$^\spadesuit$~\cite{guo2025omniaid} & ICML 2026 & 82.64 & 99.73 & 99.59 & 80.95 & 99.33 & 98.92 & 91.98 & \second{99.14} & 94.03 \\
    PGC$^\spadesuit$~\cite{PGCzhou26h} & ICML 2026 & \best{99.93} & \best{99.98} & \best{99.99} & \best{99.98} & \best{99.95} & \best{99.99} & \best{99.94} & 97.23 & \best{99.62} \\
    IAPL$^\spadesuit$~\cite{li2026towardsIAPL} & CVPR 2026 & 95.74 & 99.90 & 99.78 & 85.48 & 96.07 & 99.74 & 98.78 & 98.73 & 96.78 \\
    \midrule
    \rowcolor{oursrow}
    PE-SMP (Ours) & -- & 96.94 & \second{99.98} & \second{99.90} & \second{92.98} & \second{99.72} & \second{99.95} & \second{99.81} & \best{99.72} & \second{98.62} \\
    \bottomrule
  \end{tabular}%
  }
\end{minipage}
\par\bigskip
\begin{minipage}[t]{\textwidth}
  \centering
  \caption{\textbf{Average Precision (\%) on GenImage~\cite{zhu2023genimage}.}
    Each row uses the same checkpoint as in \cref{tab:cross_bench_ap}.
    $^\clubsuit$ denotes retraining with the official code on SDv1.4, and $^\spadesuit$ denotes evaluation with publicly released weights.
    The best and second-best results are in \best{bold} and \second{underlined}, respectively, based on unrounded scores.}
  \label{tab:genimage_ap}
  \setlength{\tabcolsep}{8pt}
  \resizebox{\textwidth}{!}{%
  \begin{tabular}{lc|cccccccc|c}
    \toprule
    Methods & Venues & Midjourney & SDv1.4 & SDv1.5 & ADM & GLIDE & Wukong & VQDM & BigGAN & Avg. \\
    \midrule
    LGrad$^\clubsuit$~\cite{tan2023LGrad} & CVPR 2023 & 85.10 & 97.85 & 97.81 & 57.25 & 80.93 & 96.04 & 46.59 & 75.98 & 79.69 \\
    NPR$^\spadesuit$~\cite{tan2024rethinkingNPR} & CVPR 2024 & 97.60 & 98.77 & 98.82 & 93.18 & 98.34 & 98.10 & 93.08 & 88.65 & 95.82 \\
    DRCT$^\spadesuit$~\cite{chen2024drct} & ICML 2024 & 99.72 & 100.00 & 99.99 & 94.47 & 97.34 & \second{100.00} & 97.00 & 98.05 & 98.32 \\
    RINE$^\clubsuit$~\cite{koutlis2024RINE} & ECCV 2024 & 99.24 & \second{100.00} & 99.99 & 97.67 & 98.46 & 100.00 & 99.75 & 97.22 & 99.04 \\
    OMAT$^\spadesuit$~\cite{zhou2026breakingOMAT} & NeurIPS 2025 & 97.05 & 100.00 & 100.00 & 93.93 & 99.94 & 100.00 & 98.85 & 99.71 & 98.68 \\
    SAFE$^\clubsuit$~\cite{li2025improvingSAFE} & KDD 2025 & 97.84 & 99.87 & 99.80 & 82.28 & 98.91 & 99.41 & 96.90 & 85.66 & 95.08 \\
    AIDE$^\spadesuit$~\cite{yan2025sanityAIDE} & ICLR 2025 & 97.95 & 99.96 & 99.98 & 94.59 & 99.08 & 99.97 & 97.07 & 93.48 & 97.76 \\
    Effort$^\spadesuit$~\cite{yan2025effort} & ICML 2025 & 98.63 & 99.99 & \second{100.00} & 99.00 & 99.74 & 99.91 & 99.60 & 98.71 & 99.45 \\
    C2P-CLIP$^\spadesuit$~\cite{tan2025c2p} & AAAI 2025 & 99.52 & 99.96 & 99.93 & 97.38 & 98.77 & 99.72 & 98.85 & 98.15 & 99.04 \\
    VIB-Net$^\spadesuit$~\cite{zhang2025towardsVIB} & CVPR 2025 & 98.46 & 99.99 & 99.95 & 92.74 & 94.96 & 99.89 & 97.87 & 84.90 & 96.09 \\
    OmniAID$^\spadesuit$~\cite{guo2025omniaid} & ICML 2026 & 96.87 & 99.99 & 99.97 & 93.23 & 99.91 & 99.94 & 98.74 & \second{99.93} & 98.57 \\
    PGC$^\spadesuit$~\cite{PGCzhou26h} & ICML 2026 & \best{100.00} & \best{100.00} & \best{100.00} & \best{100.00} & \best{100.00} & \best{100.00} & \best{100.00} & 99.92 & \best{99.99} \\
    IAPL$^\spadesuit$~\cite{li2026towardsIAPL} & CVPR 2026 & 98.97 & 99.96 & 99.87 & 98.26 & 99.36 & 99.86 & 99.70 & 99.73 & 99.46 \\
    \midrule
    \rowcolor{oursrow}
    PE-SMP (Ours) & -- & \second{99.75} & \best{100.00} & 99.96 & \second{99.26} & \second{100.00} & \best{100.00} & \second{100.00} & \best{100.00} & \second{99.87} \\
    \bottomrule
  \end{tabular}%
  }
\end{minipage}
\end{table*}

\paragraph{AIGI-Holmes.}
\Cref{tab:holmes_acc,tab:holmes_ap} report per-generator Acc and AP on
AIGI-Holmes. PE-SMP performs well on both metrics across diffusion and
autoregressive generators. On Janus and the Janus-Pro variants, PGC
achieves 91.21--96.72\% AP with near-chance Acc (50.04--50.76\%) at its
evaluation threshold, showing that good score ranking can coexist with
poor classification at a fixed threshold. PE-SMP classifies images by
the sign of its routing-weighted angular score, which indicates whether
the adapted feature is closer to the fake or real prototypes in
aggregate. This rule maintains at least 99.97\% Acc on all Janus
variants without tuning the threshold for each generator.

\begin{table*}[!t]
\begin{minipage}[t]{\textwidth}
  \centering
  \caption{\textbf{Accuracy (\%) on AIGI-Holmes~\cite{zhou2025holmes}.}
    $^\clubsuit$: retrained using the official code under the same training setting.
    $^\spadesuit$: evaluated using the publicly released weights.
    \textbf{Following the GenImage protocol, all models are trained solely on the SDv1.4 subset.}
    The best and second-best results are in \best{bold} and \second{underlined}, respectively.}
  \label{tab:holmes_acc}
  \setlength{\tabcolsep}{8pt}
  \resizebox{\textwidth}{!}{%
  \tabcolsep=8 pt
  \begin{tabular}{ll|cccccccccc|c}
    \toprule
    Methods & Venues & FLUX & Infinity & Janus & \tworow{Janus-}{Pro-1B} & \tworow{Janus-}{Pro-7B} & LlamaGen & \tworow{PixArt-}{XL} & \tworow{SD3.5-}{L} & Show-o & VAR & Avg. \\
    \midrule
    LGrad$^\clubsuit$~\cite{tan2023LGrad} & CVPR 2023 & 80.20 & 83.06 & 44.40 & 44.07 & 43.53 & 75.86 & 80.35 & 73.02 & 77.26 & 64.31 & 66.61 \\
    NPR$^\spadesuit$~\cite{tan2024rethinkingNPR} & CVPR 2024 & 90.87 & 92.80 & 60.48 & 81.02 & \second{87.69} & 92.66 & 92.08 & 88.31 & 92.53 & 89.42 & 86.79 \\
    DRCT$^\spadesuit$~\cite{chen2024drct} & ICML 2024 & 77.24 & 99.16 & 61.56 & 65.18 & 59.70 & 72.13 & 96.54 & 83.03 & 76.97 & 68.03 & 75.95 \\
    RINE$^\clubsuit$~\cite{koutlis2024RINE} & ECCV 2024 & 77.74 & 99.48 & 50.97 & 58.44 & 54.71 & 94.85 & 93.20 & 90.20 & 94.19 & 77.79 & 79.16 \\
    OMAT$^\spadesuit$~\cite{zhou2026breakingOMAT} & NeurIPS 2025 & 94.73 & 95.45 & 65.08 & 75.67 & 64.07 & 96.68 & 96.92 & 95.77 & 96.96 & 95.83 & 87.72 \\
    SAFE$^\clubsuit$~\cite{li2025improvingSAFE} & KDD 2025 & 89.14 & 99.06 & 50.43 & 50.92 & 50.38 & 81.03 & 98.69 & 84.00 & 75.21 & 75.77 & 75.46 \\
    AIDE$^\spadesuit$~\cite{yan2025sanityAIDE} & ICLR 2025 & 85.32 & 95.30 & 52.01 & 57.48 & 56.72 & 94.65 & 89.89 & 95.92 & 90.19 & 88.05 & 80.55 \\
    Effort$^\spadesuit$~\cite{yan2025effort} & ICML 2025 & 96.00 & 96.03 & 53.99 & 65.93 & 58.79 & 97.17 & 92.10 & 86.72 & 92.44 & 94.15 & 83.33 \\
    C2P-CLIP$^\spadesuit$~\cite{tan2025c2p} & AAAI 2025 & 96.12 & 99.11 & \second{72.41} & \second{86.91} & 86.04 & 98.82 & 98.89 & 98.60 & 99.02 & 93.97 & \second{92.99} \\
    VIB-Net$^\spadesuit$~\cite{zhang2025towardsVIB} & CVPR 2025 & 81.88 & 99.16 & 72.28 & 83.44 & 80.81 & 97.19 & 90.17 & 93.25 & 92.25 & 71.21 & 86.16 \\
    OmniAID$^\spadesuit$~\cite{guo2025omniaid} & ICML 2026 & 88.06 & 98.68 & 55.84 & 62.73 & 54.48 & 97.53 & 96.84 & 74.21 & 87.01 & 93.79 & 80.92 \\
    PGC$^\spadesuit$~\cite{PGCzhou26h} & ICML 2026 & \best{99.96} & \second{99.98} & 50.76 & 50.71 & 50.04 & \second{99.97} & \best{100.00} & \best{99.99} & \best{99.98} & \best{99.98} & 85.14 \\
    IAPL$^\spadesuit$~\cite{li2026towardsIAPL} & CVPR 2026 & 68.36 & 69.22 & 46.66 & 64.04 & 59.33 & 68.74 & 68.25 & 68.56 & 69.05 & 66.69 & 64.89 \\
    \midrule
    \rowcolor{oursrow}
    PE-SMP (Ours) & -- & \second{98.26} & \best{99.98} & \best{99.98} & \best{99.99} & \best{99.97} & \best{99.97} & \second{99.98} & \second{98.62} & \second{99.96} & \second{99.79} & \best{99.65} \\
    \bottomrule
  \end{tabular}%
  }
\end{minipage}
\par\bigskip
\begin{minipage}[t]{\textwidth}
  \centering
  \caption{\textbf{Average Precision (\%) on AIGI-Holmes~\cite{zhou2025holmes}.}
    $^\clubsuit$: retrained using the official code under the same training setting.
    $^\spadesuit$: evaluated using the publicly released weights.
    \textbf{Following the GenImage protocol, all models are trained solely on the SDv1.4 subset.}
    The best and second-best results are in \best{bold} and \second{underlined}, respectively.}
  \label{tab:holmes_ap}
  \setlength{\tabcolsep}{8pt}
  \resizebox{\textwidth}{!}{%
  \begin{tabular}{ll|cccccccccc|c}
    \toprule
    Methods & Venues & FLUX & Infinity & Janus & \tworow{Janus-}{Pro-1B} & \tworow{Janus-}{Pro-7B} & LlamaGen & \tworow{PixArt-}{XL} & \tworow{SD3.5-}{L} & Show-o & VAR & Avg. \\
    \midrule
    LGrad$^\clubsuit$~\cite{tan2023LGrad} & CVPR 2023 & 91.81 & 94.46 & 42.73 & 41.63 & 41.81 & 85.61 & 91.56 & 83.64 & 88.06 & 65.43 & 72.67 \\
    NPR$^\spadesuit$~\cite{tan2024rethinkingNPR} & CVPR 2024 & 97.95 & 99.13 & 63.97 & 86.07 & 90.30 & 96.91 & 98.53 & 95.34 & 98.89 & 89.82 & 91.69 \\
    DRCT$^\spadesuit$~\cite{chen2024drct} & ICML 2024 & 96.98 & 99.98 & \second{95.30} & 96.59 & 92.41 & 93.83 & 99.85 & 99.19 & 95.58 & 92.13 & 96.19 \\
    RINE$^\clubsuit$~\cite{koutlis2024RINE} & ECCV 2024 & 99.39 & 100.00 & 94.23 & \second{98.92} & \second{98.87} & 99.99 & 99.94 & \second{99.98} & 99.97 & 99.67 & \second{99.10} \\
    OMAT$^\spadesuit$~\cite{zhou2026breakingOMAT} & NeurIPS 2025 & 98.99 & 99.22 & 82.49 & 90.18 & 81.68 & 99.47 & 99.71 & 99.28 & 99.78 & 99.36 & 95.02 \\
    SAFE$^\clubsuit$~\cite{li2025improvingSAFE} & KDD 2025 & 98.85 & 99.91 & 54.07 & 68.78 & 69.35 & 95.92 & 99.89 & 97.77 & 96.52 & 94.35 & 87.54 \\
    AIDE$^\spadesuit$~\cite{yan2025sanityAIDE} & ICLR 2025 & 99.26 & 99.92 & 87.61 & 97.06 & 95.82 & 99.83 & 99.54 & 99.80 & 99.55 & 98.46 & 97.68 \\
    Effort$^\spadesuit$~\cite{yan2025effort} & ICML 2025 & 99.57 & 99.59 & 81.72 & 93.53 & 90.07 & 99.76 & 99.07 & 98.22 & 99.20 & 99.29 & 96.00 \\
    C2P-CLIP$^\spadesuit$~\cite{tan2025c2p} & AAAI 2025 & 99.50 & 99.96 & 94.22 & 97.71 & 97.85 & 99.93 & 99.96 & 99.90 & 99.85 & 98.89 & 98.78 \\
    VIB-Net$^\spadesuit$~\cite{zhang2025towardsVIB} & CVPR 2025 & 96.73 & 99.90 & 94.99 & 98.35 & 98.24 & 99.78 & 98.68 & 99.34 & 99.05 & 90.26 & 97.53 \\
    OmniAID$^\spadesuit$~\cite{guo2025omniaid} & ICML 2026 & 98.24 & 99.79 & 71.66 & 80.41 & 68.02 & 99.73 & 99.60 & 92.03 & 97.76 & 98.12 & 90.54 \\
    PGC$^\spadesuit$~\cite{PGCzhou26h} & ICML 2026 & \best{100.00} & \second{100.00} & 95.07 & 96.72 & 91.21 & \second{100.00} & \second{100.00} & \best{100.00} & \second{100.00} & \best{100.00} & 98.30 \\
    IAPL$^\spadesuit$~\cite{li2026towardsIAPL} & CVPR 2026 & 58.48 & 64.26 & 42.89 & 52.28 & 47.90 & 79.53 & 70.99 & 65.92 & 68.06 & 77.27 & 62.76 \\
    \midrule
    \rowcolor{oursrow}
    PE-SMP (Ours) & -- & \second{99.96} & \best{100.00} & \best{100.00} & \best{100.00} & \best{100.00} & \best{100.00} & \best{100.00} & 99.97 & \best{100.00} & \second{99.99} & \best{99.99} \\
    \bottomrule
  \end{tabular}%
  }
\end{minipage}
\end{table*}

\paragraph{AIGI-Now.}
\Cref{tab:aiginow,tab:aiginow_ap} report Acc and AP separately on the
pixel-level and semantic-level subsets of each generator. Every
baseline has lower mean Acc on the semantic-level subsets than on the
pixel-level subsets. For example, C2P-CLIP's mean Acc drops from
89.57\% on pixel-level subsets to 75.90\% on semantic-level subsets.
PE-SMP's mean Acc rises from 94.58\% to 97.38\%, with gains on eight of
the nine generators. The semantic-level subsets combine surreal prompts
with post-processing that disrupts pixel statistics. Across generators,
PE-SMP achieves 99.84\% mean AP on these subsets and 99.64\% on the
pixel-level subsets. Its semantic-level AP is at least 99.19\% for every
generator. SMP scores the adapted pooled PE feature against real and
fake prototype pairs, weighting each pair by the similarity of its
visual context to the input.
The semantic-level results are consistent with the score \mbox{capturing}
content differences that survive post-processing. The pixel-level
FLUX Kontext subset yields PE-SMP's lowest Acc (85.65\%). Its AP of
97.88\% indicates that the angular score still ranks images well even
when classification at the fixed threshold is less reliable.

\begin{table*}[!t]
  \centering
  \caption{\textbf{Accuracy (\%) on AIGI-Now~\cite{chen2025taskAIGInow}.}
    P and S denote the pixel-level and semantic-level subsets of each generator, respectively.
    $^\clubsuit$: retrained using the official code under the same training setting.
    $^\spadesuit$: evaluated using the publicly released weights.
    \textbf{Following the GenImage protocol, all models are trained solely on the SDv1.4 subset.}
    The best and second-best results are in \best{bold} and \second{underlined}, respectively.}
  \label{tab:aiginow}
  \setlength{\tabcolsep}{2.5pt}
  \resizebox{\textwidth}{!}{%
  \begin{tabular}{lc|cc cc cc cc cc cc cc cc cc|c}
    \toprule
    \multirow{2}{*}{Methods} & \multirow{2}{*}{Venues}
      & \multicolumn{2}{c}{Nano Banana}
      & \multicolumn{2}{c}{GPT-4o}
      & \multicolumn{2}{c}{Jimeng}
      & \multicolumn{2}{c}{Kling}
      & \multicolumn{2}{c}{Minimax}
      & \multicolumn{2}{c}{FLUX Pro}
      & \multicolumn{2}{c}{FLUX Krea}
      & \multicolumn{2}{c}{FLUX Dev}
      & \multicolumn{2}{c|}{FLUX Kontext}
      & \multirow{2}{*}{Avg.} \\
    \cmidrule(lr){3-4} \cmidrule(lr){5-6} \cmidrule(lr){7-8} \cmidrule(lr){9-10} \cmidrule(lr){11-12} \cmidrule(lr){13-14} \cmidrule(lr){15-16} \cmidrule(lr){17-18} \cmidrule(lr){19-20}
      & & P & S & P & S & P & S & P & S & P & S & P & S & P & S & P & S & P & S & \\
    \midrule
    LGrad$^\clubsuit$~\cite{tan2023LGrad} & CVPR 2023
      & 65.95 & 58.80 & 77.05 & 64.10 & 49.90 & 67.20 & 72.25 & 67.10 & 73.15 & 63.45
      & 52.00 & 66.40 & 67.10 & 59.30 & 78.80 & 64.90 & 65.50 & 56.40 & 64.96 \\
    NPR$^\spadesuit$~\cite{tan2024rethinkingNPR} & CVPR 2024
      & 88.90 & 56.40 & 90.75 & 55.90 & 77.05 & 58.20 & 89.60 & 57.00 & \second{77.00} & 57.15
      & 68.50 & 55.40 & 78.65 & 54.35 & 91.90 & 58.80 & 91.25 & 58.45 & 70.29 \\
    DRCT$^\spadesuit$~\cite{chen2024drct} & ICML 2024
      & 85.85 & 53.15 & 62.50 & 54.15 & 55.75 & 50.65 & 69.75 & 53.65 & 53.10 & 53.00
      & 56.05 & 53.15 & 54.35 & 52.95 & 78.20 & 50.40 & 66.90 & 50.45 & 58.56 \\
    RINE$^\clubsuit$~\cite{koutlis2024RINE} & ECCV 2024
      & 88.10 & 49.10 & 73.55 & 48.30 & 51.10 & 48.05 & 75.70 & 50.75 & 52.30 & 50.40
      & 53.75 & 52.00 & 52.85 & 51.70 & 88.50 & 51.10 & 80.85 & 65.55 & 60.20 \\
    OMAT$^\spadesuit$~\cite{zhou2026breakingOMAT} & NeurIPS 2025
      & 89.00 & 46.80 & 74.30 & 45.70 & 49.05 & 46.50 & 93.55 & 50.00 & 69.65 & 46.65
      & 59.00 & 51.35 & 64.80 & 46.80 & 91.10 & 47.50 & 84.75 & 50.70 & 61.51 \\
    SAFE$^\clubsuit$~\cite{li2025improvingSAFE} & KDD 2025
      & 93.80 & 54.70 & \second{99.30} & 44.70 & 50.15 & 53.30 & 86.55 & 49.80 & 50.65 & 52.80
      & 50.40 & 52.45 & 50.45 & 48.15 & 90.20 & 52.20 & 91.20 & 73.60 & 63.58 \\
    AIDE$^\spadesuit$~\cite{yan2025sanityAIDE} & ICLR 2025
      & 96.30 & 57.20 & 81.15 & 55.20 & 60.00 & 53.75 & 90.15 & 58.30 & 52.90 & 57.75
      & 57.95 & 58.20 & 51.80 & 59.65 & 89.20 & 60.10 & 88.90 & 63.60 & 66.23 \\
    Effort$^\spadesuit$~\cite{yan2025effort} & ICML 2025
      & 97.65 & 50.55 & 73.70 & 50.10 & 56.25 & 50.60 & 95.50 & 51.00 & 70.30 & 51.70
      & 56.20 & 51.55 & 76.50 & 50.45 & 88.10 & 50.20 & 68.70 & 50.55 & 63.31 \\
    C2P-CLIP$^\spadesuit$~\cite{tan2025c2p} & AAAI 2025
      & \second{98.70} & \second{77.00} & 98.70 & \second{71.85} & \second{86.90} & \second{76.20} & \second{98.70} & \second{77.45} & 63.10 & \second{76.15}
      & \second{79.50} & \second{75.85} & \second{87.85} & \second{76.15} & 98.30 & \second{76.15} & \second{94.40} & 76.30 & \second{82.74} \\
    VIB-Net$^\spadesuit$~\cite{zhang2025towardsVIB} & CVPR 2025
      & 90.75 & 68.10 & 90.55 & 68.80 & 79.30 & 61.80 & 86.05 & 69.05 & 75.25 & 69.60
      & 75.85 & 73.40 & 89.95 & 71.90 & 87.95 & 71.05 & 79.60 & 67.85 & 76.49 \\
    OmniAID$^\spadesuit$~\cite{guo2025omniaid} & ICML 2026
      & 77.10 & 50.45 & 57.75 & 48.95 & 57.95 & 50.50 & 75.60 & 49.90 & 64.75 & 49.45
      & 66.60 & 50.45 & 53.35 & 49.55 & 82.65 & 50.55 & 75.95 & 50.55 & 59.00 \\
    PGC$^\spadesuit$~\cite{PGCzhou26h} & ICML 2026
      & \best{100.00} & 50.00 & \best{100.00} & 50.40 & 50.00 & 50.10 & \best{100.00} & 50.35 & 50.15 & 50.15
      & 50.10 & 50.05 & 50.10 & 50.15 & \best{100.00} & 50.05 & \best{99.90} & \best{95.65} & 66.51 \\
    IAPL$^\spadesuit$~\cite{li2026towardsIAPL} & CVPR 2026
      & 69.15 & 50.80 & 66.70 & 56.10 & 40.55 & 47.85 & 66.95 & 62.55 & 66.75 & 53.35
      & 59.10 & 58.60 & 69.15 & 60.25 & 67.90 & 54.75 & 67.85 & 60.50 & 59.94 \\
    \midrule
    \rowcolor{oursrow}
    PE-SMP (Ours) & --
      & 94.05 & \best{98.60} & 97.05 & \best{99.95} & \best{98.05} & \best{99.60} & 91.20 & \best{96.30} & \best{91.95} & \best{94.50}
      & \best{96.60} & \best{98.50} & \best{97.40} & \best{96.55} & \second{99.30} & \best{99.80} & 85.65 & \second{92.60} & \best{95.98} \\
    \bottomrule
  \end{tabular}%
  }
\end{table*}

\begin{table*}[!t]
  \centering
  \caption{\textbf{Average Precision (\%) on AIGI-Now~\cite{chen2025taskAIGInow}.}
    P and S denote the pixel-level and semantic-level subsets of each generator, respectively.
    $^\clubsuit$: retrained using the official code under the same training setting.
    $^\spadesuit$: evaluated using the publicly released weights.
    \textbf{Following the GenImage protocol, all models are trained solely on the SDv1.4 subset.}
    The best and second-best results are in \best{bold} and \second{underlined}, respectively.}
  \label{tab:aiginow_ap}
  \setlength{\tabcolsep}{2.5pt}
  \resizebox{\textwidth}{!}{%
  \begin{tabular}{lc|cc cc cc cc cc cc cc cc cc|c}
    \toprule
    \multirow{2}{*}{Methods} & \multirow{2}{*}{Venues}
      & \multicolumn{2}{c}{Nano Banana}
      & \multicolumn{2}{c}{GPT-4o}
      & \multicolumn{2}{c}{Jimeng}
      & \multicolumn{2}{c}{Kling}
      & \multicolumn{2}{c}{Minimax}
      & \multicolumn{2}{c}{FLUX Pro}
      & \multicolumn{2}{c}{FLUX Krea}
      & \multicolumn{2}{c}{FLUX Dev}
      & \multicolumn{2}{c|}{FLUX Kontext}
      & \multirow{2}{*}{Avg.} \\
    \cmidrule(lr){3-4} \cmidrule(lr){5-6} \cmidrule(lr){7-8} \cmidrule(lr){9-10} \cmidrule(lr){11-12} \cmidrule(lr){13-14} \cmidrule(lr){15-16} \cmidrule(lr){17-18} \cmidrule(lr){19-20}
      & & P & S & P & S & P & S & P & S & P & S & P & S & P & S & P & S & P & S & \\
    \midrule
    LGrad$^\clubsuit$~\cite{tan2023LGrad} & CVPR 2023
      & 71.52 & 61.32 & 85.66 & 67.90 & 49.23 & 71.28 & 81.55 & 69.36 & 81.11 & 65.75
      & 51.49 & 69.35 & 74.09 & 60.75 & 90.36 & 68.20 & 72.29 & 58.18 & 69.41 \\
    NPR$^\spadesuit$~\cite{tan2024rethinkingNPR} & CVPR 2024
      & 95.10 & 58.71 & 98.16 & 62.74 & 82.42 & 68.32 & 97.22 & 66.74 & 83.93 & 63.56
      & 73.04 & 64.87 & 83.27 & 63.45 & 99.15 & 76.84 & 99.09 & 79.30 & 78.66 \\
    DRCT$^\spadesuit$~\cite{chen2024drct} & ICML 2024
      & 98.56 & 57.88 & 88.35 & 67.37 & 96.49 & 56.64 & 91.00 & 64.28 & 87.58 & 62.75
      & 91.79 & 60.44 & 93.42 & 61.06 & 98.24 & 56.78 & 96.59 & 54.34 & 76.86 \\
    RINE$^\clubsuit$~\cite{koutlis2024RINE} & ECCV 2024
      & 99.94 & 53.04 & 99.82 & 47.68 & 97.34 & 45.49 & 99.67 & 58.56 & \second{96.33} & 53.48
      & 95.95 & 60.19 & \second{98.74} & 60.53 & 99.87 & 59.65 & \second{99.63} & 84.38 & 78.35 \\
    OMAT$^\spadesuit$~\cite{zhou2026breakingOMAT} & NeurIPS 2025
      & 96.34 & 51.83 & 89.27 & 40.18 & 51.75 & 44.58 & 98.46 & 54.01 & 85.05 & 47.07
      & 76.72 & 56.98 & 81.47 & 46.69 & 97.17 & 50.32 & 94.09 & 50.98 & 67.39 \\
    SAFE$^\clubsuit$~\cite{li2025improvingSAFE} & KDD 2025
      & 99.57 & 55.56 & \second{99.96} & 44.52 & 66.93 & 54.96 & 98.69 & 50.31 & 68.43 & 55.10
      & 57.46 & 54.85 & 70.91 & 47.42 & 99.11 & 53.13 & 98.79 & 82.77 & 69.92 \\
    AIDE$^\spadesuit$~\cite{yan2025sanityAIDE} & ICLR 2025
      & 99.95 & \second{93.30} & 99.14 & \second{89.61} & 96.43 & 86.67 & 99.53 & \second{92.33} & 82.64 & \second{90.71}
      & 96.76 & \second{92.46} & 87.41 & \second{94.59} & 99.45 & \second{94.22} & 99.22 & 92.66 & \second{93.73} \\
    Effort$^\spadesuit$~\cite{yan2025effort} & ICML 2025
      & 99.80 & 88.55 & 96.17 & 79.71 & 87.98 & 81.53 & 99.53 & 87.93 & 95.11 & 87.72
      & 91.18 & 87.35 & 96.14 & 89.11 & 98.53 & 85.31 & 90.23 & 78.82 & 90.04 \\
    C2P-CLIP$^\spadesuit$~\cite{tan2025c2p} & AAAI 2025
      & \second{99.96} & 90.45 & 99.93 & 83.93 & \second{98.30} & \second{90.97} & \second{99.94} & 91.99 & 91.72 & 89.76
      & \second{97.10} & 90.03 & 98.03 & 90.55 & 99.75 & 89.14 & 99.06 & 82.35 & 93.50 \\
    VIB-Net$^\spadesuit$~\cite{zhang2025towardsVIB} & CVPR 2025
      & 99.05 & 77.00 & 99.20 & 76.48 & 98.41 & 70.17 & 98.24 & 77.63 & 96.58 & 78.32
      & 97.58 & 82.42 & 98.79 & 81.40 & 98.55 & 80.30 & 97.37 & 78.22 & 88.10 \\
    OmniAID$^\spadesuit$~\cite{guo2025omniaid} & ICML 2026
      & 96.99 & 72.30 & 88.03 & 62.81 & 75.80 & 66.59 & 96.01 & 67.76 & 88.95 & 61.60
      & 87.58 & 66.34 & 77.16 & 60.06 & 95.94 & 63.79 & 92.91 & 60.85 & 76.75 \\
    PGC$^\spadesuit$~\cite{PGCzhou26h} & ICML 2026
      & \best{100.00} & 79.14 & \best{100.00} & 84.47 & 75.07 & 79.94 & \best{100.00} & 84.90 & 76.97 & 74.85
      & 79.04 & 78.97 & 79.84 & 77.28 & \best{100.00} & 78.39 & \best{99.99} & \second{98.91} & 85.99 \\
    IAPL$^\spadesuit$~\cite{li2026towardsIAPL} & CVPR 2026
      & 55.04 & 61.24 & 53.38 & 68.96 & 39.44 & 61.96 & 53.19 & 70.65 & 55.13 & 61.24
      & 52.48 & 71.54 & 58.60 & 64.91 & 54.50 & 64.34 & 63.03 & 66.84 & 59.81 \\
    \midrule
    \rowcolor{oursrow}
    PE-SMP (Ours) & --
      & 99.78 & \best{99.97} & 99.91 & \best{99.99} & \best{99.96} & \best{100.00} & 99.59 & \best{99.83} & \best{99.74} & \best{99.68}
      & \best{99.92} & \best{99.95} & \best{99.97} & \best{99.92} & \second{99.98} & \best{100.00} & 97.88 & \best{99.19} & \best{99.74} \\
    \bottomrule
  \end{tabular}%
  }
\end{table*}

\subsection{UniversalFakeDetect setting}

\Cref{tab:ufd_ap} reports AP on the 19 UniversalFakeDetect test subsets,
complementing the Acc results in Table~3 of the main paper. After
training on four ProGAN categories, PE-SMP ranks second to IAPL on both
mean metrics, with 94.65\% mAcc and 98.86\% mAP, trailing IAPL by 0.96 and
0.46 percentage points, respectively. AP exceeds 99\% on every GAN,
LDM, and GLIDE subset. Acc also remains high on the LDM and GLIDE
subsets, showing that both the score ranking and fixed-threshold
decisions transfer from the ProGAN training source to diffusion outputs.

The image restoration subsets SITD and SAN are more challenging for
PE-SMP. It obtains 79.17\% Acc and 91.63\% AP on SITD, and 67.12\% Acc
and 92.24\% AP on SAN. The AP on SAN suggests that the score still
provides useful ranking information on this subset.

SITD and SAN contain images restored from real inputs, so the input and
output can depict the same scene and differ mainly in fine appearance
details. SMP scores pooled features against context-specific prototype
pairs. When these features describe nearly identical scene content,
small restoration artifacts may produce little angular separation
between the classes. This is consistent with the weaker ranking and
the particularly low fixed-threshold Acc on these subsets.

\begin{table*}[!t]
  \centering
  \caption{\textbf{Evaluation on the UniversalFakeDetect dataset~\cite{ojha2023towards} using Average Precision (\%).}
    Results of the other methods are cited from IAPL~\cite{li2026towardsIAPL}.
    $^\clubsuit$: retrained using the official code under the same training setting.
    \textbf{The 4-class ProGAN is used as the training data.}
    The best and second-best results are in \best{bold} and \second{underlined}, respectively.}
  \label{tab:ufd_ap}
  \setlength{\tabcolsep}{2.2pt}
  \resizebox{\textwidth}{!}{%
  \begin{tabular}{l|cccccc c cc cc c ccc ccc c|c}
    \toprule
    \multirow{2}{*}{Methods}
      & \multicolumn{6}{c}{GAN}
      & \multirow{2}{*}{\tworow{Deep}{fakes}}
      & \multicolumn{2}{c}{Low level}
      & \multicolumn{2}{c}{Perceptual loss}
      & \multirow{2}{*}{Guided}
      & \multicolumn{3}{c}{LDM}
      & \multicolumn{3}{c}{Glide}
      & \multirow{2}{*}{Dalle}
      & \multirow{2}{*}{Avg.} \\
    \cmidrule(lr){2-7} \cmidrule(lr){9-10} \cmidrule(lr){11-12} \cmidrule(lr){14-16} \cmidrule(lr){17-19}
      & \tworow{Pro-}{GAN} & \tworow{Cycle-}{GAN} & \tworow{Big-}{GAN}
      & \tworow{Style-}{GAN} & \tworow{Gau-}{GAN} & \tworow{Star-}{GAN}
      &
      & SITD & SAN
      & CRN & IMLE
      &
      & \tworow{200}{steps} & \tworow{200}{w/cfg} & \tworow{100}{steps}
      & \tworow{100}{27} & \tworow{50}{27} & \tworow{100}{10}
      & & \\
    \midrule
    FatFormer~\cite{liu2024fatformer}
      & 100.00 & 100.00 & \best{99.98} & 99.75 & 100.00 & 100.00
      & 97.99 & 97.94 & 81.21 & 99.84 & 99.93 & 91.99
      & 99.81 & 99.09 & 99.87 & 99.13 & 99.41 & 99.20 & 99.82 & 98.16 \\
    FreqNet~\cite{tan2024frequencyFreqNet}
      & 99.92 & 99.63 & 96.05 & 99.89 & 99.71 & 98.63
      & \best{99.92} & 94.42 & 74.59 & 80.10 & 75.70 & 96.27
      & 96.06 & \best{100.00} & 62.34 & 99.80 & 99.78 & 96.39 & 77.78 & 91.95 \\
    NPR~\cite{tan2024rethinkingNPR}
      & 100.00 & 99.53 & 94.53 & 99.94 & 88.82 & 100.00
      & 84.41 & \second{97.95} & \best{99.99} & 50.16 & 50.16 & 98.26
      & 99.92 & 99.91 & 99.92 & \best{99.87} & \best{99.89} & \best{99.92} & 99.26 & 92.76 \\
    RINE~\cite{koutlis2024RINE}
      & 100.00 & 100.00 & 99.90 & 99.40 & 100.00 & 100.00
      & 97.90 & 97.20 & 94.90 & 97.30 & 99.70 & 96.40
      & 99.80 & 98.30 & 99.90 & 98.80 & 99.30 & 98.90 & 99.30 & 98.79 \\
    AIDE~\cite{yan2025sanityAIDE}
      & 100.00 & 99.89 & 94.44 & \best{100.00} & 97.69 & 100.00
      & 76.35 & 83.60 & 88.82 & 90.73 & 99.94 & 97.28
      & 99.29 & 99.12 & 99.34 & 99.30 & 99.25 & 99.19 & 99.03 & 95.96 \\
    C2P-CLIP~\cite{tan2025c2p}
      & 100.00 & 100.00 & 99.96 & 99.50 & 100.00 & 100.00
      & \second{98.59} & \best{98.92} & 84.56 & 99.86 & 99.95 & 94.13
      & \best{99.99} & 99.83 & \best{99.98} & 99.72 & 99.79 & 99.83 & \second{99.91} & 98.66 \\
    MiraGe~\cite{shi2025mirage}
      & 100.00 & \second{100.00} & 99.90 & 99.80 & 99.90 & 99.90
      & 96.00 & 93.90 & 84.70 & 99.90 & \second{100.00} & 96.40
      & 99.90 & 99.10 & 99.90 & 99.80 & 99.70 & 99.80 & 99.90 & 98.34 \\
    SAFE~\cite{li2025improvingSAFE}
      & 100.00 & 99.77 & 95.88 & 99.93 & 97.19 & 100.00
      & 97.45 & 85.07 & \second{99.27} & 41.32 & 47.23 & 95.81
      & \second{99.96} & \second{99.92} & \second{99.96} & 98.93 & 99.21 & 99.36 & 99.67 & 92.42 \\
    Effort$^\clubsuit$~\cite{yan2025effort}
      & 100.00 & \best{100.00} & \second{99.97} & 98.54 & 100.00 & \second{100.00}
      & 95.25 & 71.14 & 89.70 & 99.53 & 99.94 & 87.13
      & 99.86 & 97.98 & 99.86 & 98.51 & 98.99 & 98.52 & 99.66 & 96.56 \\
    PGC~\cite{PGCzhou26h}
      & 99.80 & 99.90 & 99.70 & 94.80 & 100.00 & 97.60
      & 94.30 & 84.90 & 99.00 & 96.50 & 99.30 & \best{99.60}
      & 98.70 & 99.90 & 99.80 & 99.60 & 99.60 & 99.70 & 99.70 & 98.02 \\
    IAPL~\cite{li2026towardsIAPL}
      & \second{100.00} & 99.99 & 99.95 & 99.75 & \second{100.00} & \best{100.00}
      & 97.59 & 97.27 & 98.12 & \second{99.97} & \best{100.00} & 96.25
      & 99.86 & 99.60 & 99.74 & 99.40 & 99.73 & 99.86 & \best{99.95} & \best{99.32} \\
    \midrule
    \rowcolor{oursrow}
    PE-SMP (Ours)
      & \best{100.00} & 99.64 & 99.94 & \second{99.98} & \best{100.00} & 99.89
      & 97.90 & 91.63 & 92.24 & \best{99.98} & 99.99 & \second{98.28}
      & 99.94 & 99.62 & 99.95 & \second{99.85} & \second{99.86} & \second{99.87} & 99.76 & \second{98.86} \\
    \bottomrule
  \end{tabular}%
  }
\end{table*}

\begin{table*}[!t]
  \centering
  \caption{Backbone comparison with a linear head and the SMP head on the same input. Accuracy (\%) on seven generalization benchmarks. The PE-G rows use block-50 patch features, and the other rows use last-block patch features. The PE-G configuration with the SMP head is PE-SMP. Subscripts give the change of the SMP head over the linear head on the same backbone.}
  \label{tab:ablation_backbone_heads}
  \setlength{\tabcolsep}{5pt}
  \resizebox{\textwidth}{!}{%
  \begin{tabular}{l|ll|ccccccc|c}
    \toprule
    Backbone & Scale & Head & GenImage & Chameleon & CommunityAI & SocialRF & WildRF & AIGI-Holmes & AIGI-Now & Avg. \\
    \midrule
    CLIP & L & Linear & 75.10 & 45.91 & 49.81 & 72.43 & 76.05 & 79.53 & 74.24 & 67.58 \\
    CLIP & L & SMP & 79.61\up{4.51} & 61.56\up{15.65} & 60.57\up{10.76} & 70.52\down{1.91} & 73.88\down{2.17} & 83.50\up{3.97} & 72.62\down{1.62} & 71.75\up{4.17} \\
    \cmidrule(lr){1-11}
    DINOv3 & L & Linear & 82.99 & 59.79 & 60.83 & 59.00 & 62.53 & 83.48 & 81.42 & 70.01 \\
    DINOv3 & L & SMP & 87.61\up{4.62} & 82.34\up{22.55} & 83.62\up{22.79} & 83.18\up{24.18} & 86.69\up{24.16} & 91.39\up{7.91} & 80.23\down{1.19} & 85.01\up{15.00} \\
    \cmidrule(lr){1-11}
    PE & L & Linear & 83.85 & 80.53 & 84.06 & 95.85 & 95.06 & 88.70 & 76.55 & 86.37 \\
    PE & L & SMP & 92.00\up{8.15} & 91.44\up{10.91} & 92.92\up{8.86} & 92.95\down{2.90} & 96.67\up{1.61} & 95.34\up{6.64} & 83.93\up{7.38} & 92.18\up{5.81} \\
    \cmidrule(lr){1-11}
    DINOv3 & H+ & Linear & 84.18 & 53.58 & 53.87 & 52.23 & 52.55 & 75.70 & 74.99 & 63.87 \\
    DINOv3 & H+ & SMP & 93.06\up{8.88} & 85.33\up{31.75} & 88.08\up{34.21} & 89.88\up{37.65} & 92.23\up{39.68} & 95.07\up{19.37} & 82.84\up{7.85} & 89.50\up{25.63} \\
    \midrule
    PE & G & Linear & 96.53 & 71.54 & 78.53 & 84.15 & 91.22 & 98.00 & \textbf{96.56} & 88.08 \\
    \rowcolor{oursrow}
    PE & G & SMP & \textbf{98.62}\up{2.09} & \textbf{96.07}\up{24.53} & \textbf{96.66}\up{18.13} & \textbf{95.03}\up{10.88} & \textbf{98.29}\up{7.07} & \textbf{99.65}\up{1.65} & 95.98\down{0.58} & \textbf{97.19}\up{9.11} \\
    \bottomrule
  \end{tabular}}
\end{table*}

\twocolumn[{%
  \centering
  \includegraphics[width=\textwidth]{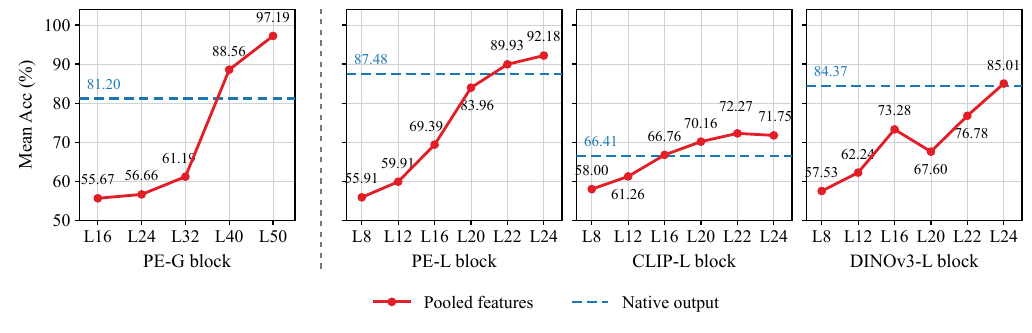}
  \captionof{figure}{\textbf{Effect of feature depth across frozen backbones.}
    Within each backbone, we apply the same readout procedure at every
    tested block to obtain the pooled feature $q$ for each image. We report
    mean accuracy across the seven benchmarks in the GenImage setting.
    The horizontal blue dashed lines mark the results using
    each backbone's native output. The vertical dashed line separates
    PE-G from the three L models, which are evaluated at the same six
    block indices.}
  \label{fig:ablation_feature_depth}
  \par
}]

\raggedbottom
\section{Additional ablations}

We train the ablation variants on the GenImage SD v1.4 split. The
backbone and head comparisons and the ablations on feature depth,
PE-SMP components, and unit normalization cover all seven benchmarks in
the GenImage setting. The two hyperparameter sweeps evaluate the
residual angle cap and the number of contexts on Chameleon.

\paragraph{Head comparison across frozen backbones.}
\Cref{tab:ablation_backbone_heads} compares a linear head with SMP
using the same frozen features within each backbone configuration.
SMP improves mean Acc in all five configurations. DINOv3-H+ shows the
largest gain in mean Acc, at 25.63 percentage points (63.87\% to
89.50\%). Its Acc on WildRF rises by 39.68 points (52.55\% to 92.23\%).
On PE-G, the largest gains are on Chameleon and CommunityAI, at 24.53
and 18.13 points, respectively. On AIGI-Now, PE-G accuracy changes from
96.56\% with the linear head to 95.98\% with SMP.

SMP combines a shared real-to-fake offset with a local offset for each
visual context to form a pair of real and fake prototypes. Soft routing
weights the angular score from each pair according to the input's
visual context. The gains on WildRF, Chameleon, and CommunityAI are
consistent with the benefit of scoring varied in-the-wild images
against relevant prototype pairs. The gains in mean Acc across CLIP,
DINOv3, and PE suggest that this design can use visual context across
different pretrained feature spaces.

\paragraph{Feature depth across frozen backbones.}
\Cref{fig:ablation_feature_depth} shows mean Acc with SMP using pooled
features from different blocks of each frozen backbone. The blue
dashed lines show the results using the corresponding native outputs.
In the PE-G panel, mean Acc rises from 61.19\% at block 32 to 88.56\% at
block 40 and reaches its highest tested value at block 50. The largest
increase occurs in the later blocks of the network. We therefore sample
PE-L, CLIP-L, and DINOv3-L every four blocks from block 8 to 20, then
every two blocks through block 24. PE-L and DINOv3-L achieve their
highest tested mean Acc at block 24. The denser sampling captures
CLIP-L's best tested result at block 22 (72.27\%), followed by a slight
drop at block 24 (71.75\%). For each backbone, the best tested block
outperforms its native output.


\paragraph{Discriminative head.}
\Cref{tab:ablation_head} compares the linear, MLP, and SMP heads on
the same PE-G block-50 features. SMP achieves the highest accuracy
on six of the seven benchmarks. Its largest gains over the stronger
baseline on each benchmark are on Chameleon and CommunityAI, at 23.38
and 18.13 percentage points, respectively.

\begin{table*}[!t]
  \centering
  \caption{Ablation on the discriminative head. All heads use the same layer-50 features. Accuracy (\%) on seven generalization benchmarks.}
  \label{tab:ablation_head}
  \setlength{\tabcolsep}{12pt}
  \resizebox{\textwidth}{!}{%
  \tabcolsep=12 pt
  \begin{tabular}{l|ccccccc|c}
    \toprule
    Head & GenImage & Chameleon & CommunityAI & SocialRF & WildRF & AIGI-Holmes & AIGI-Now & Avg. \\
    \midrule
    w/ Linear head & 96.53 & 71.54 & 78.53 & 84.15 & 91.22 & 98.00 & 96.56 & 88.08 \\
    w/ MLP head & 98.44 & 72.69 & 76.83 & 81.65 & 90.57 & 99.22 & \textbf{98.36} & 88.25 \\
    \midrule
    \rowcolor{oursrow}
    PE-SMP & \textbf{98.62} & \textbf{96.07} & \textbf{96.66} & \textbf{95.03} & \textbf{98.29} & \textbf{99.65} & 95.98 & \textbf{97.19} \\
    \bottomrule
  \end{tabular}}
\end{table*}

\begin{table*}[!t]
  \centering
  \caption{Ablation of PE-SMP. Accuracy (\%) on seven generalization benchmarks.
    Without the residual MLP, the bounded adapter is disabled.
    Without prototype training, the prototype pairs remain fixed at their
    source-data initialization, and only the bounded adapter is trained.
    The variant without soft routing uses hard routing to select one
    prototype pair per image. The variant without block 50 uses the native
    PE-G output as input to SMP.}
  \label{tab:ablation_components}
  \setlength{\tabcolsep}{8.48pt}
  \resizebox{\textwidth}{!}{%
  \begin{tabular}{l|ccccccc|c}
    \toprule
    Method & GenImage & Chameleon & CommunityAI & SocialRF & WildRF & AIGI-Holmes & AIGI-Now & Avg \\
    \midrule
    w/o residual MLP & 97.74 & 85.41 & 90.48 & \textbf{95.83} & \textbf{98.32} & 99.33 & \textbf{96.26} & 94.77 \\
    w/o prototype training & 97.90 & 87.09 & 91.35 & 95.38 & 98.30 & 99.30 & 95.64 & 95.00 \\
    w/o soft routing & 97.20 & 83.38 & 88.84 & 93.12 & 96.22 & 99.00 & 96.18 & 93.42 \\
    w/o block 50 & 96.59 & 70.85 & 73.84 & 74.17 & 82.12 & 91.27 & 79.57 & 81.20 \\
    \midrule
    \rowcolor{oursrow}
    PE-SMP & \textbf{98.62} & \textbf{96.07} & \textbf{96.66} & 95.03 & 98.29 & \textbf{99.65} & 95.98 & \textbf{97.19} \\
    \bottomrule
  \end{tabular}}
\end{table*}

\begin{table}[!t]
  \centering
  \caption{Effect of using the unit sphere. PE-SMP uses unit-normalized features and prototypes. The three ablations omit prototype normalization, feature normalization, and both, respectively, in the distance computation.
    mAcc and mAP are averaged over seven benchmarks. RAcc and FAcc denote
    accuracy on real and generated images, respectively.}
  \label{tab:ablation_unit_norm_main}
  \resizebox{\linewidth}{!}{%
  \setlength{\tabcolsep}{13pt}
  \begin{tabular}{lcccc}
    \toprule
    Method & mAcc & mAP & RAcc & FAcc \\
    \midrule
    w/o proto. norm & 95.55 & 99.61 & 93.58 & 97.51 \\
    w/o feat. norm & 94.92 & 99.65 & 91.84 & \textbf{98.00} \\
    w/o both & 95.06 & 99.60 & 92.47 & 97.66 \\
    \midrule
    \rowcolor{oursrow}
    PE-SMP & \textbf{97.19} & \textbf{99.71} & \textbf{96.42} & 97.95 \\
    \bottomrule
  \end{tabular}}
\end{table}

\paragraph{Ablation of PE-SMP.}
\Cref{tab:ablation_components} isolates the block-50 feature readout,
the bounded adapter, learned prototypes, and soft routing. Using the
native PE-G output lowers accuracy on all seven benchmarks. Among the changes to the head, hard routing causes the
largest drop in mean accuracy (3.77 points). On Chameleon, removing the bounded
adapter and freezing the prototypes lower accuracy to 85.41\% and
87.09\%, respectively.

The variant without the bounded adapter achieves slightly higher Acc on
SocialRF (95.83\% vs.\ 95.03\%) and WildRF (98.32\% vs.\ 98.29\%). This
suggests that the unadapted feature $q$ already supports effective
prototype scoring on these benchmarks, where the adjustment learned on
the source split may offer limited benefit. Hard routing lowers Acc on
both benchmarks. Across all seven benchmarks, PE-SMP achieves the
highest mean Acc (97.19\% vs.\ 95.00\% for the strongest ablation).
Each head ablation lowers mean Acc, supporting the use of the bounded
adapter, prototype learning, and soft routing together.

\paragraph{Effect of using the unit sphere.}
To evaluate the effect of using the unit sphere, we compare PE-SMP,
which uses unit-normalized features and prototypes, with variants that
reintroduce their original norms into the distance computation,
separately or jointly. In these variants, we use a distance based on
vector lengths and the angle between them,
\begin{equation}
  D(r, s, \theta) = \sqrt{(r - s)^2 + r s\,\theta^2},
\end{equation}
which reduces to $\theta$ when $r = s = 1$. Here $r$ denotes the feature
norm and $s$ the prototype norm. The feature-norm ablation uses the
original feature norm for $r$ and sets $s=1$. The prototype-norm
ablation uses the original prototype norm for $s$ and sets $r=1$.
The joint ablation uses both original norms. All variants use the same
routing, bounded adapter, prototype initialization, losses, parameter
count, and training schedule as PE-SMP.

Table~\ref{tab:ablation_unit_norm_main} shows that mAcc falls from
97.19\% with PE-SMP to 95.55\% without prototype normalization and
94.92\% without feature normalization. Removing both yields 95.06\%,
close to removing feature normalization alone. The drop is concentrated
on real images. RAcc falls from 96.42\% to 91.84--93.58\%, while FAcc
remains near 98\%. Across all configurations, mAP stays within
99.60--99.71\%, suggesting that normalization mainly improves decisions
at the fixed threshold, with little change in ranking quality.

In Table~\ref{tab:ablation_unit_norm}, the largest drops occur on
Chameleon and CommunityAI, where PE-SMP achieves 96.07\% and 96.66\%,
respectively. The ablated variants reach 86.89--88.62\% on Chameleon
and 91.47--93.10\% on CommunityAI. Their accuracy remains close to
PE-SMP on GenImage and AIGI-Holmes. Some ablated variants achieve
slightly higher Acc on individual benchmarks. PE-SMP achieves the
highest mAcc across all seven benchmarks, supporting the use of the
unit sphere for classification across domains at a fixed threshold.

\begin{table*}[!t]
  \centering
  \caption{Effect of using the unit sphere across benchmarks.
    Per-benchmark accuracy (\%) for the normalization variants in
    Table~\ref{tab:ablation_unit_norm_main}. mAcc is the mean accuracy
    across the seven benchmarks.}
  \label{tab:ablation_unit_norm}
  \setlength{\tabcolsep}{9pt}
  \resizebox{\textwidth}{!}{%
  \begin{tabular}{l|ccccccc|c}
    \toprule
    Variant & GenImage & Chameleon & CommunityAI & SocialRF & WildRF & AIGI-Holmes & AIGI-Now & mAcc \\
    \midrule
    w/o prototype norm & 97.52 & 88.62 & 93.10 & \textbf{95.93} & \textbf{98.36} & 99.26 & 96.03 & 95.55 \\
    w/o feature norm & 97.66 & 86.89 & 91.47 & 94.43 & 97.61 & 99.40 & \textbf{96.98} & 94.92 \\
    w/o both & 97.62 & 87.13 & 91.79 & 95.32 & 98.02 & 99.28 & 96.26 & 95.06 \\
    \midrule
    \rowcolor{oursrow}
    PE-SMP & \textbf{98.62} & \textbf{96.07} & \textbf{96.66} & 95.03 & 98.29 & \textbf{99.65} & 95.98 & \textbf{97.19} \\
    \bottomrule
  \end{tabular}}
\end{table*}

\paragraph{Residual angle cap on Chameleon.}
\Cref{tab:ablation_angle_cap_chameleon_split_mixed} shows that the
$30^\circ$ cap gives the highest Acc and AP. A tight cap may limit useful
feature corrections. A loose cap lets the adapted feature move farther
from the original feature $q$ used for routing. Loosening or removing the
cap mainly reduces real-image accuracy, with little change in AP.

\paragraph{Number of contexts on Chameleon.}
\Cref{tab:ablation_num_contexts_chameleon_split} shows that $K=16$
achieves the highest Acc and AP. With too few contexts, each prototype
pair may need to cover a wider range of image content. With many
contexts, the fixed initialization set is divided among more clusters,
leaving fewer examples per pair on average. The largest drops in Acc
come mainly from labeling more real images as fake.

\begin{table*}[!t]
  \begin{minipage}[t]{.48\textwidth}
    \centering
    \caption{Ablation on the angle cap of the bounded residual on Chameleon. Accuracy (\%) on real and fake images, their average (Acc), and AP (\%).}
    \label{tab:ablation_angle_cap_chameleon_split_mixed}
    \resizebox{\linewidth}{!}{%
    \tabcolsep=12 pt
    \begin{tabular}{lcccc}
      \toprule
       & \multicolumn{4}{c}{Chameleon} \\
      \cmidrule(lr){2-5}
      Angle cap & Real & Fake & Acc & AP \\
      \midrule
      $10^\circ$ & 88.45 & 98.84 & 93.65 & 99.39 \\
      $20^\circ$ & 76.67 & 99.33 & 88.00 & 99.38 \\
      \rowcolor{oursrow}
      $30^\circ$ (PE-SMP) & 93.20 & 98.94 & \textbf{96.07} & \textbf{99.51} \\
      $45^\circ$ & 71.10 & 99.46 & 85.28 & 99.29 \\
      $60^\circ$ & 59.87 & 99.53 & 79.70 & 99.14 \\
      Unbounded & 64.11 & 99.57 & 81.84 & 99.26 \\
      \bottomrule
    \end{tabular}
    }
  \end{minipage}%
  \hfill
  \begin{minipage}[t]{.48\textwidth}
    \centering
    \caption{Ablation on the number of contexts $K$ on Chameleon. Accuracy (\%) on real and fake images, their average (Acc), and AP (\%).}
    \label{tab:ablation_num_contexts_chameleon_split}
    \resizebox{\linewidth}{!}{%
    \tabcolsep=13 pt
    \begin{tabular}{lcccc}
      \toprule
       & \multicolumn{4}{c}{Chameleon} \\
      \cmidrule(lr){2-5}
      $K$ & Real & Fake & Acc & AP \\
      \midrule
      1  & 78.58 & 99.27 & 88.93 & 99.35 \\
      2  & 89.69 & 98.38 & 94.04 & 98.85 \\
      4  & 27.42 & 99.87 & 63.64 & 98.86 \\
      8  & 98.49 & 93.16 & 95.82 & 98.99 \\
      \rowcolor{oursrow}
      16 (PE-SMP) & 93.20 & 98.94 & \textbf{96.07} & \textbf{99.51} \\
      32 & 82.15 & 99.22 & 90.69 & 99.39 \\
      64 & 48.02 & 99.73 & 73.87 & 99.18 \\
      \bottomrule
    \end{tabular}
    }
  \end{minipage}
\end{table*}

\begin{figure*}[!t]
  \centering
  \includegraphics[width=\textwidth]{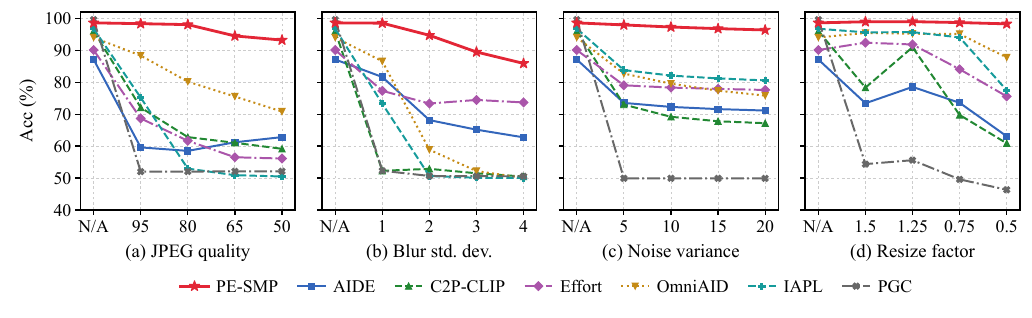}
  \caption{\textbf{Detection accuracy under four image perturbations on GenImage.}
    Panels (a)--(d) show JPEG compression, Gaussian blur, Gaussian noise,
    and resizing, respectively. N/A denotes unperturbed images and
    corresponds to a resize factor of $1\times$.
    Noise variance is measured on the $[0,255]$ pixel scale, and resizing
    uses bilinear interpolation.}
  \label{fig:robustness_full}
\end{figure*}

\newpage
\section{Additional robustness results}

\Cref{fig:robustness_full} compares seven detectors trained on GenImage
SD v1.4. Each perturbation is applied separately at test time before
the detector's usual input preprocessing. We test JPEG quality factors
of 95, 80, 65, and 50, and Gaussian blur standard deviations of 1, 2,
3, and 4 pixels. We also evaluate additive zero-mean Gaussian noise
with variances of 5, 10, 15, and 20 on the $[0,255]$ pixel scale, and
bilinear resizing at scale factors of $1.5$, $1.25$, $0.75$, and $0.5$.
We average Acc equally across GenImage test subsets. PE-SMP leads in
every perturbed setting. Its Acc stays above 98\% under both upsampling
and downsampling, and declines only modestly as Gaussian noise variance
increases. JPEG compression and blur have a larger effect at stronger
settings, where fine image details are attenuated. Blur causes the
largest decrease in Acc, with PE-SMP retaining 85.9\% at the strongest
tested level. These results suggest that PE-SMP's prototype-based
scores continue to distinguish real and generated images after common
post-processing.

\clearpage

\flushbottom

\end{document}